\documentclass{article}

\usepackage[left=1in, right=1in]{geometry}
\usepackage{amsmath,amssymb}
\usepackage{graphicx}

\usepackage{microtype}
\usepackage{graphicx}
\usepackage{subfigure}
\usepackage{booktabs} 

\usepackage{subcaption} 
\usepackage{graphicx} 
\usepackage{float} 
\usepackage{graphicx, subcaption}
\usepackage{stfloats} 
\usepackage{hyperref}

\usepackage{listings}

\usepackage{amsmath}
\usepackage{amssymb}
\usepackage{mathtools}
\usepackage{amsthm}

\DeclareMathOperator*{\argmin}{arg\,min}

\usepackage[capitalize,noabbrev]{cleveref}
\theoremstyle{plain}
\newtheorem{theorem}{Theorem}[section]

\theoremstyle{definition}

\theoremstyle{remark}

\usepackage[textsize=tiny]{todonotes}

\usepackage{xcolor}

\usepackage{algorithm}
\usepackage{algorithmic}

\usepackage{inputenc}

\definecolor{vpurple}{RGB}{161,0,255}
\renewcommand*{\mathcolor}{}
\def\mathcolor#1#{\mathcoloraux{#1}}
\newcommand*{\mathcoloraux}[3]{%
  \protect\leavevmode
  \begingroup
    \color#1{#2}#3%
  \endgroup
}

\usepackage{hyperref}       
\usepackage{amsthm}         

\usepackage{bm}  

\theoremstyle{plain}

\usepackage{optidef}

\begin{document}

\title{ReFill: Reinforcement Learning for Fill-In Minimization}

\author{%
  Elfarouk Harb\\
  University of Illinois Urbana-Champaign\\
  \texttt{eyfmharb@gmail.com} \\ 
  \and 
  Ho Shan Lam\\
  The Trade Desk\\
  \texttt{sharonhslhk@gmail.com}
}
\date{}
\maketitle

\begin{abstract}
Efficiently solving sparse linear systems \( \mathbf{A x} = \mathbf{b} \), where \( \mathbf{A} \) is a large, sparse, symmetric positive semi-definite matrix, is a core challenge in scientific computing, machine learning, and optimization. A major bottleneck in Gaussian elimination for these systems is \textit{fill-in}—the creation of non-zero entries that increase memory and computational cost. Minimizing fill-in is NP-hard, and existing heuristics like Minimum Degree and Nested Dissection offer limited adaptability across diverse problem instances.

We introduce \textit{ReFill}, a reinforcement learning framework enhanced by Graph Neural Networks (GNNs) to learn adaptive ordering strategies for fill-in minimization. ReFill trains a GNN-based heuristic to predict efficient elimination orders, outperforming traditional heuristics by dynamically adapting to the structure of input matrices. Experiments demonstrate that ReFill outperforms strong heuristics in reducing fill-in, highlighting the untapped potential of learning-based methods for this well-studied classical problem.
\end{abstract}

\section{Introduction and Background}
In this paper, we consider the fundamental problem of solving the system of equations ${\bf A x} = {\bf b}$, where ${\bf A}$ is a $n\times n$ symmetric positive semi-definite matrix. Such sparse linear systems show up naturally in scientific computing, machine learning, theoretical computer science, and scalable optimization \cite{crespelle2023survey, brezinski2022journey, bliznets2020lower, bollhofer2020state, georgenesteddissection, rosetarjan, ROSE1970597, ROSE1972183}. These systems are instrumental in numerous domains ranging from finite element analysis to large graph-based models in machine learning. Typically, Gaussian elimination serves as a primary solution method. However, the efficiency of Gaussian elimination is highly sensitive to matrix ordering, as different variable elimination orderings introduce varying levels of \textit{fill-in}—the additional non-zero entries created during variable elimination, which directly impact memory usage and runtime. If the matrix ${\bf A}$ is dense, then Gaussian Elimination requires $\Theta(n^3)$ time. However, if ${\bf A}$ is sparse, we might be able to save time by avoiding explicit manipulation of zeros or fill-in. Reducing fill-in, therefore, is critical for optimizing resource use and enabling scalable computation in high-dimensional settings.

\begin{figure}
    \centering
    \includegraphics[width=0.45\linewidth]{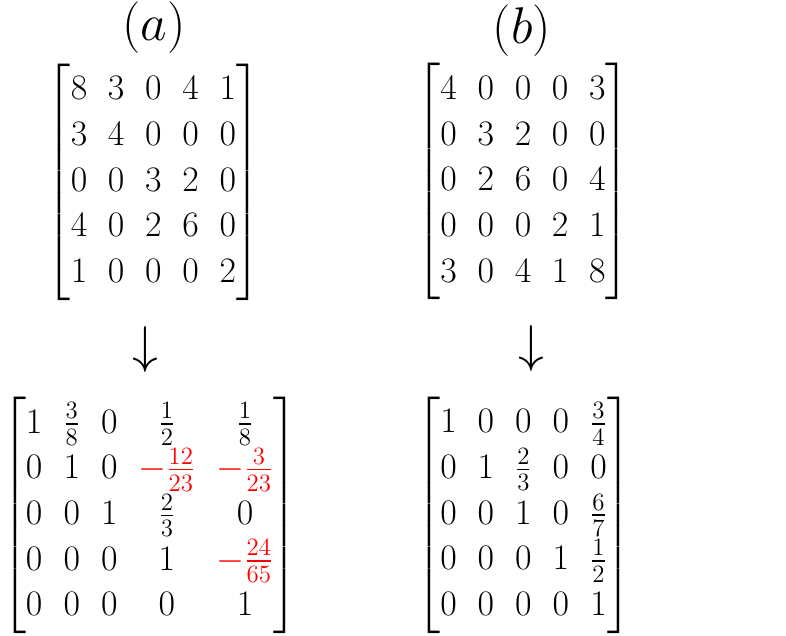}
\caption{\textbf{(a)} A symmetric positive-definite matrix \( \mathbf{A} \) and its factorization after Gaussian elimination with a fixed variable elimination order. This ordering introduces three fill-in entries at positions \( A_{2,4} \), \( A_{2,5} \), and \( A_{4,5} \), which were initially zero but became non-zero during elimination. \textbf{(b)} The same matrix after applying a permutation \( \pi = (5,1,2,3,4) \) via permutation matrix \( \mathbf{P}_{\pi} \), which reorders the rows and columns of \( \mathbf{A} \) by moving the first row to the last row, and the first column to the last position resulting in matrix \( \mathbf{A}' \). This reordering leads to a matrix where Gaussian elimination introduces no fill-in, resulting in lower memory usage and faster computation.}
\label{fig:elimination_order_1}
\end{figure}

One way to minimize fill-in during Gaussian elimination is to reorder the rows and columns of ${\bf A}$ to solve an equivalent system \(\mathbf{PAP}^T\), where \(\mathbf{P}\) denotes a permutation matrix. See \cref{fig:elimination_order_1} for an example where different variable elimination order in Gaussian elimination leads to fewer fill-in.  This approach, introduced in the seminal works by Tarjan and Rose \cite{ROSE1970597, ROSE1972183, rosetarjan}, reinterprets fill-in into a graph-theoretic perspective, thereby providing structural insights into sparse matrices. Specifically, we will recap  \textit{elimination orders} and \textit{fill-in minimization} within the context of graph theory.

Let us consider a symmetric positive semi-definite system \(\mathbf{A x} = \mathbf{b}\), where the sparse structure of \(\mathbf{A}\) can be represented by an undirected graph \(G(\mathbf{A})\). In this graph, each vertex corresponds to a row (or column) in \(\mathbf{A}\), with an edge \(ij, i\neq j\) present if \(\mathbf{A}_{i,j} \neq 0\). The process of Gaussian elimination on \(\mathbf{A}\) can then be interpreted in terms of \textit{eliminating} vertices in \(G(\mathbf{A})\). Specifically, eliminating a vertex \(i\) in \(G'\) entails connecting all neighbors of \(i\) (denoted \(\mathcal{N}_{G'}(i)\)) to form a clique, followed by the deletion of \(i\) from \(G'\). This in turn adds ``fill-in'' edges, neighbors of $i$ that were previously not connected. 

\begin{figure*}
    \centering
    \includegraphics[width=0.45\linewidth]{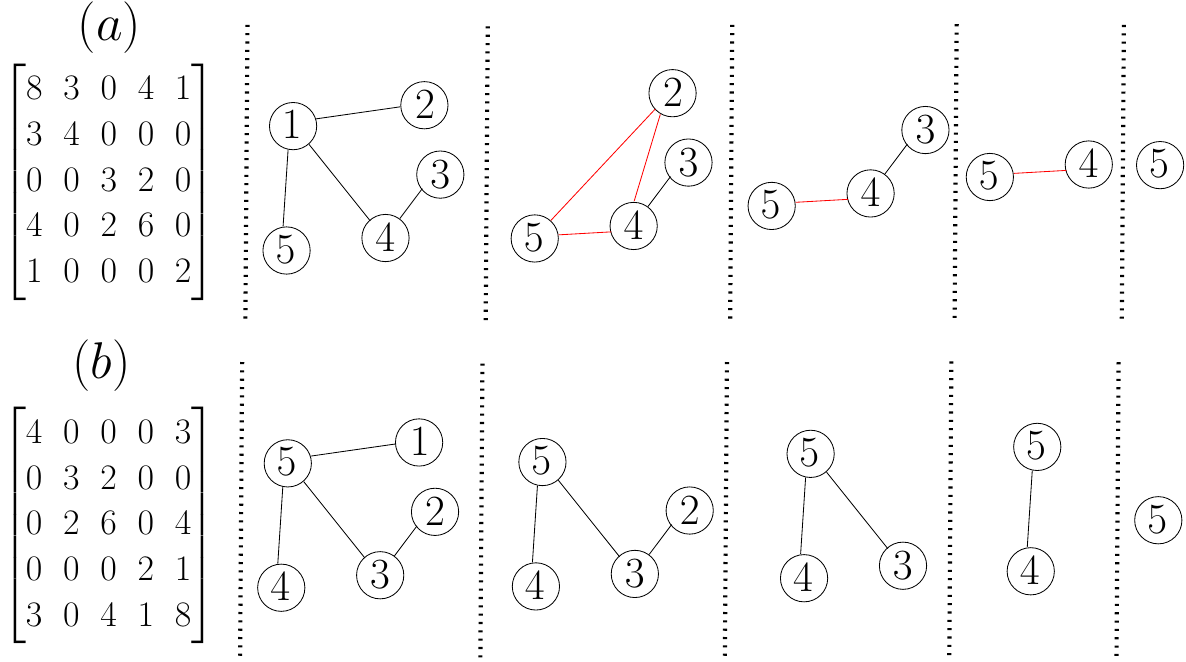}
\caption{\textbf{(a)} The matrix \( \mathbf{A} \) from \cref{fig:elimination_order_1} and its corresponding graph representation \( G(\mathbf{A}) \). Vertices are eliminated in the order \( 1,2,3,4,5 \). Eliminating vertex 1 forces its neighbors \( (2,4,5) \) to form a clique, introducing three fill-in edges \( (2,4), (2,5), (4,5) \) (shown in red) before vertex 1 is removed. These fill-in edges directly correspond to the non-zero entries added during Gaussian elimination in \cref{fig:elimination_order_1}. \textbf{(b)} The permuted matrix \( \mathbf{A}' \) and its associated graph \( G(\mathbf{A}') \), obtained by reordering the rows and columns of \( \mathbf{A} \) by moving the first row to be the last row, and the first column to be the last column (i.e. according to the permutation $\pi=(51234)$). Although the graph is structurally identical to \( G(\mathbf{A}) \), the new vertex ordering eliminates all variables without introducing any fill-in edges, showing how proper ordering prevents unnecessary fill-in.}
    \label{fig:EliminationOrder2}
\end{figure*}

Each permutation matrix \(\mathbf{P}\) determines a specific ordering \(\pi\) of the vertices. Executing Gaussian elimination according to this ordering \(\pi = (\pi(1), \ldots, \pi(n))\) means sequentially eliminating variables in the specified order. The edges that emerge from this process are collectively known as the \textit{fill-in}. See \cref{fig:EliminationOrder2} for an illustration. 

To formalize this process, consider the iterative graph sequence associated with an ordering \(\pi\):
\begin{enumerate}
    \item Set \(G^{(0)} \leftarrow G(\mathbf{A})\).
    \item For each \(i = 1\) to \(n\):
        \begin{itemize}
            \item Define \(\mathcal{P}_{\pi}(i) =  \{ (v, w) : v,w \in \mathcal{N}_{G^{(i-1)}}(\pi(i)), v\neq w, (v, w)\not\in E(G^{i-1}) \} \) as the fill-in edges that will be added by eliminating $\pi(i)$. 
            \item Eliminate \(\pi(i)\) from \(G^{(i-1)}\), resulting in the updated graph \(G^{(i)} = G^{(i-1)}\cup \mathcal{P}_{\pi}(i)\setminus \{\pi(i)\}  \).
        \end{itemize}
\end{enumerate}

The edges \(E_{\pi} = \cup_i \mathcal{P}_\pi(i)\) thus represent the fill-in that occurs when performing Gaussian elimination in the order \(\pi\).

The following theorem offers a characterization of the fill-in associated with any ordering \(\pi\).

\begin{theorem}[\cite{rosetarjan}]
\label{thm:1}
    Given an ordering \(\pi\) and assuming no ``lucky cancellations''\footnote{Lucky cancellations happens for some matrices ${\bf A}$ where some non-zero elements incidentally get cancelled to zeros when eliminating a variable. In practice, this is unlikely to happen, so one often ignores their effect.}, an edge \(ij \in E_{\pi}\) exists if and only if \(i \neq j\) and there exists a path \(v_0 = i, v_1, \dots, v_k, v_{k+1} = j\) in \(G(\mathbf{A})\) satisfying
    \[
    \max_{1 \leq r \leq k} \pi(v_r) < \min(\pi(i), \pi(j)).
    \]
\end{theorem}

Unfortunately, while we can characterize the fill-in for any given ordering \(\pi\), finding an optimal ordering \(\pi^\ast\) that minimizes the fill-in is known to be NP-Complete \cite{npcomplete}. Moreover, several inapproximation results are known. For example in \cite{inapprox}, the existence of polynomial time approximation schemes for this problem is ruled out, assuming ${\bf P} \neq {\bf NP}$, and the existence of 
a $2^{O(n^{1-\delta})}$-time approximation schemes for any positive $\delta$ is also ruled out, assuming the Exponential Time Hypothesis. Hence apriori, the problem might seems hopeless to tackle. Despite the theoretical computational difficulty of this problem, several heuristics, such as the Minimum Degree, Minimum Fill-In, and Nested Dissection heuristics, have been proposed and extensively studied in the literature \cite{markowitz, georgenesteddissection, ROSE1972183, rosetarjan}. These heuristics remain widely used in practice for fill-in minimization and are surprisingly effective in practice, often being within a few percentages off from the optimal fill-in order.

\paragraph{Minimum Degree Heuristic, \textsc{MDH}} This heuristic is a natural greedy algorithm, initially introduced by Markowitz \cite{markowitz}, and was popularized by Rose in their PhD thesis. The algorithm is simple; it creates the elimination order $\pi$ dynamically. It starts with $G^{(0)} \leftarrow G({\bf A})$. In iteration $i\geq 1$, the algorithm picks the vertex $\pi(i)$ with minimum degree (ties broken arbitrarily) in $G^{(i-1)}$:
\begin{align*}
    \pi(i) = \argmin_{u\in G^{(i-1)}} \mathrm{deg}_{G^{(i-1)}}(u)
\end{align*}

The algorithm then eliminates $\pi(i)$ (by making its neighbors a clique, adding potentially some fill-in edges $\mathcal{P}_{\pi}(i)$, then removing $\pi(i)$). Finally, it sets $G^{(i)}\leftarrow G^{(i-1)}\cup \mathcal{P}_{\pi}(i) \setminus \{\pi(i)\}$ and goes to iteration $i+1$. 

The algorithm can be implemented in $O(mn)$ time where $m, n$ are the number of edges and vertices respectively \cite{fastmindegree}. Moreover, Cummings, Fahrbach, and Fatehpuria \cite{fastmindegree} showed that under the strong exponential time hypothesis, no $O(nm^{1-\epsilon})$ time algorithm exists for any $\epsilon > 0$.  

The Minimum degree heuristic has given rise to \textit{hundreds} of research papers on improving the running time of its practical implementations \cite{tiebreaking}, and it is widely used and implemented in practice. 

\paragraph{Minimum Fill-In Heuristic, \textsc{MFillH}} This  heuristic is also another natural greedy algorithm that also creates its elimination order $\pi$ dynamically. It starts with $G^{(0)}\leftarrow G({\bf A})$. In iteration $i\geq 1$, it picks the vertex $\pi(i)$ that minimizes the fill-in in $G^{(i-1)}$ of eliminating $\pi(i)$:
\begin{align*}
\textsc{FillIn}(u) 
&= \left|\bigl\{(v,w) \mid v,w \in \mathcal{N}_{G^{(i-1)}}(u), \right.\\
&\quad \left. ~~~~(v,w)\not\in E\bigl(G^{(i-1)}\bigr)\bigr\}\right|\\
    \pi(i) & = \argmin_{u\in G^{(i-1)}} \textsc{FillIn}(u).
\end{align*}

The algorithm then eliminates $\pi(i)$. Finally, it sets $G^{(i)}\leftarrow G^{(i-1)}\cup \mathcal{P}_{\pi}(i) \setminus \{\pi(i)\}$ and goes to iteration $i+1$.

\paragraph{Importance of Tie-Breaking.} The heuristics discussed above handle ties in an arbitrary manner; specifically, when multiple vertices have the same minimum degree or fill-in value, any one of them may be selected for elimination. Some research has focused on developing more sophisticated tie-breaking rules to enhance the effectiveness of these heuristics. For example, some methods combine multiple heuristics, using one heuristic to resolve ties determined by another, which can sometimes result in improved fill-in but does not consistently guarantee better fill-in \cite{tiebreaking, multiplemdh}. Additionally, the \textit{Multiple Minimum Degree Heuristic} (MMDH) has been introduced, which involves removing all vertices that share the minimum degree simultaneously \cite{multiplemdh}. This approach strikes a balance between the aggressive removal of numerous vertices in a single step and the potential suboptimal fill-in introduced by making sequential, single-vertex decisions. By deleting multiple vertices at once, MMDH can reduce the overall computational complexity while maintaining a reasonable level of fill-in.

\paragraph{Nested Dissection}  
Nested dissection is a powerful method that leverages graph separators to recursively decompose a graph into smaller, more manageable subproblems \cite{georgenesteddissection, ROSE1972183, rosetarjan}. By identifying and removing a small ``separator'' set of vertices (or edges), the graph is partitioned into subgraphs that can be processed independently, after which the solutions are combined. This approach is especially efficient for classes of graphs where small separators are guaranteed to exist, such as planar graphs, thanks to well-known separator theorems. 

Beyond its extensive use in practice for tasks like sparse matrix factorization, nested dissection also shines in theoretical settings; notably, the fastest known algorithm for maximum matching in planar graphs is built upon the framework of nested dissection \cite{Mucha2006}, underscoring its fundamental importance in both theory and application.

\paragraph{How Do The Heuristics Compare?}  
In practice, the minimum degree heuristic (\textsc{MDH}) often runs faster and is simpler to implement, since it only needs to identify the current vertex with the smallest degree at each step. By contrast, the minimum fill-in heuristic (\textsc{MFillH}) invests more computation to estimate which vertex’s elimination would incur the fewest additional edges. While \textsc{MFillH} sometimes achieves smaller overall fill-in, this more sophisticated approach does not guarantee a strictly better ordering on every graph, and it can be significantly more expensive to run. Nested dissection, on the other hand, can deliver very effective results when a small separator is available or can be computed. However, finding such separators is not always straightforward for arbitrary graphs. Consequently, \textsc{MDH} tends to be the go-to choice for speed and ease of implementation, \textsc{MFillH} can yield sparser factorizations in select cases (albeit at greater cost), and nested dissection stands out when its underlying structural assumptions, such as the presence of small separators, are well satisfied. Regardless, all three heuristics usually get almost optimal fill-in on most graphs. 

\paragraph{Motivation and Contributions}  

Many real-world applications require solving linear systems repeatedly with identical sparsity patterns, or sparsity patterns drawn from the same distribution, such as in PDE simulations and real-time control, where even a slight improvement in fill-in reduction translates into considerable time and memory savings across \textit{multiple solves}. Traditional heuristics like Minimum Degree work well for single-shot scenarios but fail to adapt to recurring solves, leaving room for improvement, particularly when the matrix structure is complex or irregular. For these multi-solve settings, existing methods cannot dynamically refine their strategies. Nested dissection offers strong performance for grid-like problems but relies on domain-specific insights about separators, making it unsuitable for general matrix classes.

Surprisingly, despite decades of research on fill-in minimization, no existing heuristic learns from repeated interactions with the matrix structure. Methods like Minimum Degree and Minimum Fill-In treat each solve independently, missing opportunities to refine the elimination order over time. Most importantly, they are as-is, their fill-in cannot be improved. 

We address this challenge with \textsc{ReFill}, a reinforcement learning framework that combines Graph Neural Networks (GNNs) and a sequential decision-making process to dynamically learn effective elimination orders over repeated solves. Unlike traditional methods, ReFill adapts to the problem’s structure by refining its strategy as it interacts with the matrix. Our experiments show that ReFill consistently outperforms classical heuristics, offering a promising, data-driven approach to a longstanding problem in scientific computing.

\paragraph{A Bird's-Eye View of Our Method}  
At the heart of our approach is the idea of casting fill-in minimization as a reinforcement learning (RL) problem, where each sparse matrix (or equivalently its associated graph) becomes a separate ``game'', analogous to ``atari game levels''. The current graph structure, along with any vertices already eliminated, constitutes the game’s “state,” and each possible vertex elimination action yields a ``cost'' equal to the fill-in caused by eliminating that vertex. The agent attempts to learn an order dynamically, by making sequential elimination decisions at each iteration guided by a GNN.

In principle, any of the remaining vertices at a state could be eliminated at any given iteration. However, this leads to a blow up in size of the the action space which makes the sampling-complexity required to learn a ``good'' ordering high. To keep learning tractable despite the large action space, we employ a targeted action masking strategy that restricts choices to vertices that either have minimum degree or would induce minimum fill-in upon elimination. This is a combination of both the \textsc{MDH} and \textsc{MFillH} heuristics. Essentially, the agent is learning when to apply which heuristic on which graph, as well as the tie breaking rules using the Neural Network. This controlled subset of actions dramatically reduces sampling complexity necessary to learn good policies, while still preserving a high chance of including the truly optimal choice in most scenarios since these heuristics often perform exceptionally well and are near-optimal. We then train our policy network, implemented via Graph Convolutional Networks, using the Proximal Policy Optimization (PPO) algorithm, allowing it to systematically refine its choices to produce increasingly better elimination orderings. 

\paragraph{Reinforcement Learning for Combinatorial Optimization.} Reinforcement learning (RL) has emerged as a powerful paradigm for solving combinatorial optimization (CO) problems such as the Traveling Salesman Problem (TSP), knapsack, and vehicle routing. Traditional exact methods like branch-and-bound often fail to scale, while heuristic approaches require significant domain expertise. Recent work leverages RL to learn policies that iteratively construct solutions, achieving competitive results on benchmarks like the TSP \cite{kool2019attention}, Scheduling \cite{igorgnnjscheduling} and many other problems; see a recent survey at \cite{victorgnn}. Frameworks such as \textit{OR-Gym} \cite{tang2021or} standardize RL-based CO experimentation, enabling reproducible comparisons between learned policies and classical algorithms. Notably, graph neural networks (GNNs) have become instrumental in encoding combinatorial structures, with architectures like Graph Attention Networks (GATs) enabling RL agents to reason over graph-based CO problems \cite{DBLP:journals/corr/abs-1710-10903,joshi2019efficient}. For example, \cite{khalil2017learning} demonstrated that GNNs paired with RL can learn heuristics for graph traversal tasks, achieving near-optimal solutions on Maximum Cut instances with 100 nodes.

Surveys emphasize the growing adoption of RL-GNN frameworks \cite{victorgnn}, particularly for their ability to generalize across problem sizes and constraints \cite{extrmalgraphalpha, bengio2021machine,mazyavkina2021reinforcement}. Despite progress, challenges persist in reward shaping for sparse CO environments and scaling to industrial-scale instances. 

\paragraph{Organization}  
Section 2 presents a review of related work, focusing primarily on reinforcement learning, graph neural networks, and Proximal Policy Optimization (PPO). Readers already familiar with these topics may wish to skip directly to Section 3, where we formally describe our new RL-based framework for fill-in minimization and detail the underlying architecture. Section 4 provides experimental results that demonstrate the effectiveness of the proposed method, and Section 5 offers future work and limitations.

\section{Related Work}
Beside the related work on fill-in minimization discussed in the introduction, we briefly recap known results on Graph Neural Networks (GNNs), Markov Decision Processes (MDPs), and the PPO optimization algorithm. If the reader is familiar with said topics, they are encouraged to skip this section. 

\paragraph{Evolution of Graph Neural Networks.} The emergence of graph neural networks (GNNs) has revolutionized machine learning on non-Euclidean data through their ability to capture complex relational patterns \cite{DBLP:journals/tnn/WuPCLZY21, DBLP:journals/corr/abs-2003-00982, DBLP:journals/tkde/ZhangCZ22}. This paradigm shift has spawned diverse architectural families: recurrent frameworks that propagate states through iterative refinement \cite{DBLP:journals/corr/LiTBZ15, DBLP:conf/nips/NicolicioiuDL19}, spectral-spatial convolutional operators \cite{DBLP:conf/csonet/ZhangTXM18, DBLP:conf/iclr/KipfW17}, reconstruction-driven auto-encoders \cite{DBLP:journals/corr/KipfW16a, DBLP:journals/corr/abs-1802-08773}, attention-based neighborhood weighting mechanisms \cite{DBLP:journals/air/SunLLZMRW23, DBLP:journals/corr/abs-1710-10903}, and transformer-inspired architectures  \cite{DBLP:conf/nips/YunJKKK19, DBLP:journals/corr/abs-2302-04181}.

These architectures power a wide application landscape spanning structural analysis tasks like node classification \cite{DBLP:conf/bigdataconf/IzadiFSL20, DBLP:journals/corr/abs-2109-05641, DBLP:journals/corr/abs-2306-16976} and link prediction \cite{DBLP:conf/icml/ChenWHDL20, DBLP:conf/icml/ShirzadVVSS23, DBLP:conf/nips/ZhuZXT21}, to whole-graph characterization for molecular property prediction \cite{DBLP:conf/nips/RongBXX0HH20, DBLP:conf/iclr/HuLGZLPL20, DBLP:journals/corr/Altae-TranRPP16} and community detection \cite{DBLP:conf/iclr/ChenLB19, DBLP:journals/tsp/LevieMBB19, DBLP:conf/sigir/0001GRTY20}. Such versatility stems from their foundational mechanism: learnable message passing that recursively transforms node representations through localized information aggregation \cite{scarselli2008graph}.

Consider an undirected graph $G=(V,E)$ with adjacency matrix $\textbf{A} \in \{0,1\}^{|V|\times|V|}$. The message propagation dynamics can be formalized through layer-wise updates:
\[
\boldsymbol{H}^{(k+1)} = \sigma\left(\boldsymbol{\tilde{A}}\boldsymbol{H}^{(k)}\boldsymbol{W}^{(k)}\right),
\]
where $\boldsymbol{\tilde{A}} = \boldsymbol{A} + \boldsymbol{I}$ introduces self-connections, $\boldsymbol{W}^{(k)}$ represents learnable parameters, and $\sigma$ denotes nonlinear activation. This operation induces local node update rules:
\[
h_i^{(k+1)} = \sigma\left(\sum_{j \in \mathcal{N}(i)} \boldsymbol{W}^{(k)}h_j^{(k)}\right),
\]
with $\mathcal{N}(i)$ denoting node $i$'s neighborhood. Modern implementations generalize this through learnable aggregation functions:
\[
h_i^{(k+1)} = \text{MLP}^{(k)}\left(h_i^{(k)}, \bigoplus_{j \in \mathcal{N}(i)} \phi(h_i^{(k)}, h_j^{(k)}, e_{ij})\right),
\]
where $\bigoplus$ represents permutation-invariant aggregation (e.g., sum, mean, or max) and $\phi$ encodes edge features when available. This  propagation enables GNNs to develop good representations while preserving structural relationships.

\paragraph{Reinforcement Learning in Games.}
Reinforcement Learning (RL) has demonstrated remarkable success in complex game environments, wherein an autonomous agent interacts with an environment (or game) by observing states, taking actions, and receiving rewards \cite{sutton2018reinforcement, mnih2015human}.
Formally, RL problems are often framed as a Markov Decision Process (MDP), defined by a set of states $\mathcal{S}$, a set of actions $\mathcal{A}$, a transition distribution $p(s_{t+1} \mid s_{t}, a_{t})$, and a reward function $r(s_{t}, a_{t})$.
In the context of \emph{games}, the \textit{observation space} (or state space) might include raw pixels (in video games), board configurations (in board games), or symbolic features (in card games).
The \textit{action space} typically comprises valid moves or control signals (e.g. joystick commands in Atari games, or placing a stone on a board in Go).
A \textit{reward} is then provided by the environment based on the outcome of each action, for example, immediate point increments in Atari games or win/loss signals at the end of a board game \cite{silver2016mastering}.
By learning a policy that maps observations to actions with the objective of \textit{maximizing} expected cumulative reward, RL agents have achieved superhuman performance in diverse domains, such as Atari games \cite{mnih2015human} and the game of Go \cite{silver2017mastering}.

\paragraph{Proximal Policy Optimization (PPO).}
Among numerous RL algorithms, \emph{Proximal Policy Optimization (PPO)} \cite{schulman2017proximal} is a popular policy gradient method that balances sample efficiency, stability, and ease of implementation. It is often used to learn neural network policies for RL algorithms. PPO alternates between sampling data through interaction with the environment and optimizing a clipped objective function that constrains large policy updates.
Specifically, PPO uses a \emph{clipped} surrogate objective,
\[
L^{\text{CLIP}}(\theta) = \mathbb{E}_t \Big[ 
  \min \bigl(r_t(\theta)\hat{A}_t, \,\text{clip}\bigl(r_t(\theta), 1-\epsilon, 1+\epsilon\bigr)\hat{A}_t \bigr)
\Big],
\]
where $r_t(\theta) = \frac{\pi_\theta(a_t \mid s_t)}{\pi_{\theta_\mathrm{old}}(a_t \mid s_t)}$ is the probability ratio of the new policy $\pi_\theta$ to the old policy $\pi_{\theta_\mathrm{old}}$, and $\hat{A}_t$ is an estimator of the advantage function.

By clipping the probability ratio $r_t(\theta)$ to the interval $[1-\epsilon,\,1+\epsilon]$, PPO avoids excessively large policy updates, thus preventing instability or catastrophic performance collapses. This mechanism has made PPO a standard choice for many benchmark tasks in continuous control, robotics, and game-based RL environments. In this paper, we use the MaskedPPO implementation which allows action masking by sampling only from ``valid'' actions in each state \cite{maskedppo}. 

\section{ReFill}

\paragraph{Problem Statement.} The problem is modeled as a sequential decision-making task. Given the matrix \({\bf A}\) and an ordering \(\pi\) of the vertices of \(G({\bf A})\), let \(\textsc{FillInCost}(G({\bf A}), \pi)\) denote the total fill-in cost from eliminating the vertices in order \(\pi(1), \dots, \pi(n)\). The NP-hard optimization problem is 

\begin{equation}
\label{objective}
    \min_{\pi \in \mathbb{S}_n} \textsc{FillInCost}(G({\bf A}), \pi).
\end{equation}

\paragraph{Sequential Decision-Making.} Instead of learning the entire order at once, we aim to learn a policy \(\pi^\ast(u \mid G) \in (0, 1)\), which gives the probability of eliminating a vertex \(u \in V(G)\). For any graph \(G\), this policy determines the next vertex to eliminate, ensuring that the permutation \(\pi^\ast\) for Eq.~(\ref{objective}) can be recovered (breaking ties arbitrarily).

\paragraph{Supervised Learning Challenges.} Developing data-driven heuristics to outperform both the minimum degree (\textsc{MDH}) and minimum fill-in (\textsc{MFillH}) heuristics is inherently challenging due to the lack of high-quality training data. While the minimum degree heuristic is computationally efficient and often produces near-optimal results, there exist graph instances where the optimal elimination order deviates from \textsc{MDH} or \textsc{MFillH}. To train a model capable of identifying such deviations, we would require representative graphs with known optimal elimination orders that highlight scenarios where neither heuristic suffices. However, obtaining such data is fundamentally difficult, as computing the optimal solution for fill-in minimization is NP-hard. This computational barrier makes it infeasible to generate large-scale datasets of optimal or near-optimal solutions. Consequently, this lack of accessible ground truth poses a bottleneck in leveraging supervised learning approaches to learn policies that generalize beyond existing heuristics.

\paragraph{First Attempt: Reinforcement Learning Over All Actions.} One way we can tackle this problem is using reinforcement learning with a massive action and state space. The states correspond to the current graph, and the action at each step is to select a node for elimination. The reward for each action is defined as the negative of the number of fill-in edges added by eliminating the chosen node. The objective is to find an elimination sequence that maximizes the sum of rewards; effectively minimizing the total fill-in. However, the action space is combinatorially large, as it includes all possible nodes that can be selected at each step, and the state space, defined by the set of possible graphs resulting from different elimination orders, grows exponentially with the number of vertices. This makes the training very computationally expensive and intractable from a sampling-complexity point of view. It also makes it more likely that the agent can get stuck in a local optima elimination order. 

\paragraph{A Glimmer Of Hope.} While this combinatorial explosion poses a significant challenge, a glimmer of hope is that traditional heuristics like \textsc{MDH} and \textsc{MFillH}, and their tie-breaking rules, are extremely \textbf{local heuristics}. These heuristics make locally optimal choices based on simple local graph properties, such as a node's degree and its neighbors degrees. Despite this, these methods do extremely well in practice. So one would hope that a shallow GNN using the message passing mechanism of information from its ``nearby'' vertices can learn a more clever heuristic. Moreover, one would hope that this heuristic is \textbf{graph-size agnostic}. Specifically, it calculates a ``goodness'' score for each node based on it's nearby neighborhood, and finally take the node with the best goodness score with a softmax layer. 

\paragraph{\textsc{ReFill}.}
The previous discussion points to a natural solution framework. The environment is initialized with the input graph for which we are trying to find a good elimination order. At any given time, the agent evaluates the graph's current state, which is derived from the original graph by eliminating vertices. To make an elimination decision, a graph neural network (GNN) processes the node  features producing a ``goodness'' score for each node that reflects its suitability for elimination. The GNN is trained to learn these scores by leveraging local graph properties through message passing. It should also be noted that we are not explicitly deleting a node during elimination; we are simply masking over the decision of deleting it again using the deleted mask stored in the observation space. The ``game'' ends when all vertices have been deleted (i.e. the deleted mask is all ones).

The action space is constrained by masking out nodes that do not satisfy key heuristic properties, such as having the minimum degree or inducing the least fill-in. This dramatically reduces the action space's size, allowing the reinforcement learning (RL) agent to focus on decisions that are more likely to yield optimal or near-optimal results. \textit{Essentially, the agent is learning when to apply which heuristic based on the graph properties}. 

\begin{figure}[h]
    \centering
    \includegraphics[width=0.55\linewidth]{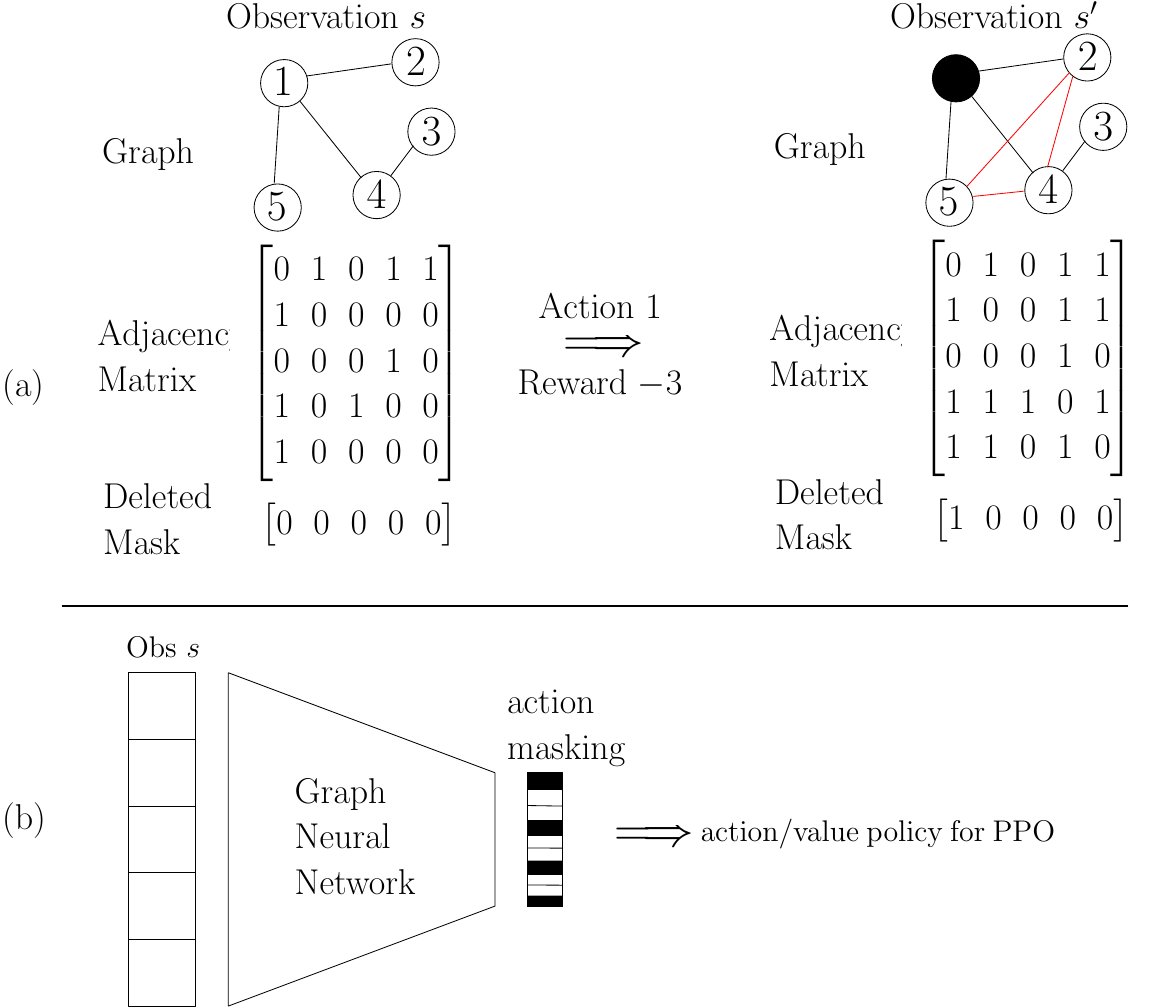}
    \caption{\textsc{ReFill} reinforcement learning environment. (a) An example observation $s$ is updated by deleting vertex 1, introducing 3 fill-in edges ($-3$ reward). (b) Schematic of the RL loop using a GNN and masked PPO.}
    \label{fig:architecture}
\end{figure}

The policy is optimized using the Proximal Policy Optimization (PPO) algorithm. PPO balances exploration and exploitation by limiting large updates to the policy. The reward signal is the negative of the fill-in edges added during the elimination step, encouraging the agent to minimize fill-in over the sequence of elimination steps.  This iterative process enables the RL agent to refine its elimination policy through repeated interactions with the graph. 
\cref{fig:architecture} summarizes the architecture and training process for \textsc{ReFill}. 

\begin{table}
\caption{ReFill Fill-in compared to \textsc{MDH} and \textsc{MFillH} on all datasets. Percentages reported are $\frac{\textsc{MDH}-\textsc{ReFill}}{\textsc{MDH}}$  and $\frac{\textsc{MFillH} - \textsc{ReFill}}{\textsc{MFillH}}$. Positive values indicate \textsc{ReFill} outperforming the heuristic.  }
\label{table1}
\vskip 0.1in
\begin{center}
\begin{small}
\begin{sc}
\begin{tabular}{lccccr}
\toprule
Graph & $V$ & $E$ & \textsc{ReFill} & \textsc{MDH} & \textsc{MFillH} \\
Grid 5x5 & 25 & 40 & 37 & 0.0\% & 0.0\%\\
Grid 6x6 & 36 & 60 & 69 & 2.8\% & 2.8\%\\
Grid 7x7 & 49 & 84 & 111 & 6.7\% & 1.8\%\\
Grid 8x8 & 64 & 112 & 166 & 9.3\% & 9.8\%\\
Grid 9x9 & 81 & 144 & 240 & 10.4\% & 0.8\%\\
Grid 10x10 & 100 & 180 & 325 & 13.6\% & 7.4\%\\
2.graph & 129 & 4943 & 195 & 2.5\% & 7.1\%\\
3.graph & 101 & 840 & 286 & 12.0\% & 0.0\%\\
11.graph & 126 & 1095 & 183 & 6.2\% & 3.2\%\\
13.graph & 119 & 161 & 92 & 0.0\% & 1.1\%\\
18.graph & 150 & 259 & 105 & 2.8\% & 5.4\%\\
23.graph & 200 & 661 & 799 & 4.9\% & 5.4\%\\
26.graph & 120 & 4904 & 229 & 9.8\% & 0.9\%\\
40.graph & 147 & 7303 & 352 & 6.6\% & 3.3\%\\
92.graph & 132 & 255 & 202 & 4.7\% & -1.5\%\\
99.graph & 166 & 396 & 382 & 18.6\% & 5.0\%\\
100.graph & 152 & 377 & 360 & 6.0\% & 1.1\%\\
\midrule
\bottomrule
\end{tabular}
\end{sc}
\end{small}
\end{center}
\vskip -0.1in
\end{table}

\section{Experiments}
In this section, we evaluate the performance of our reinforcement learning (RL)-based algorithm, \textsc{ReFill}. We compare our method against the established minimum degree heuristic (\textsc{MDH}) and the minimum fill-in heuristic (\textsc{MFillH}) on both synthetic and real-world graphs. We also present an analysis of training dynamics, fill-in cost, and briefly discuss ablation and runtime overhead.

\subsection{Experimental Setup}
\paragraph{Graph Instances.}
We evaluate our methods on the following graph classes:
\paragraph{Grid graph.} We use the $N=n\times n$ grid graph for $5\leq n \leq 10$. The $n\times n$ grid system occurs naturally when solving differential equations with finite differences method; it was the initial motivation for introducing the nested dissection heuristic~\cite{georgenesteddissection}.
\paragraph{PACE 2017 Track-B Public Dataset.} The goal of the yearly PACE challenge is to investigate the applicability of algorithmic ideas studied and developed in the subfields of multivariate, fine-grained, parameterized, or fixed-parameter tractable algorithms. The objective of the PACE 2017 Track-B challenge~\cite{pace} was to solve the NP-hard Minimum Fill-In problem \textbf{exactly} (i.e., no heuristics) within 30 minutes on real-world data. The test datasets were curated by the authors from multiple sources. For optimizing Gaussian elimination, they selected instances from the Matrix Market~\cite{matrixmarket} and the Network Repository~\cite{networkrepo}, where exact solutions could be computed. We use 11 medium-sized graphs from the public datasets, $\textsc{n.graph}$ for $N\in \{2,3,11,13,18,23,26,40,92,99,100\}$. \Cref{table1} shows the number of vertices and edges for each graph.
\paragraph{Generalization Experiment.} In addition, to test the generalization of the model, we conduct the following experiment. We sample $35$ graphs from $G(50, 0.2)$, where $G(n, p)$ is the Erdős–Rényi graph with $50$ vertices and each (undirected) edge selected with probability $0.2$. We train the GNN on all these graphs \textit{simultaneously} using action masking, then evaluate the trained heuristic on $200$ \textit{new} $G(50, 0.2)$ graphs, and report the average improvement using the learned heuristic. We emphasize this is using a \textbf{single set} of learned parameters, and not one per graph.

\paragraph{Evaluation Metrics.}
For every graph, we measure the total fill-in cost, defined as the number of missing edges that are added when deleting nodes in a particular order. We select the best elimination order found by our RL algorithm. These costs are then compared to \textsc{MDH} and \textsc{MFillH}.

\subsubsection{Implementation Details}
Our RL model uses a policy network based on a 2-layer graph convolutional network (GCN) with average aggregation. Each node has three features: (i) the normalized degree in the current graph, (ii) the fill-in that would be introduced by deleting the node, and (iii) whether the node is already eliminated. We train using PPO with masking~\cite{maskedppo} for 500,000 timesteps, where a timestep here represents the elimination of a single vertex. See \cref{ImplementationDetails} for the remaining implementation details. 

\subsection{Performance on Grid Graphs and PACE Dataset}
\Cref{table1} summarizes the total fill-in cost for \textsc{ReFill}, \textsc{MDH}, and \textsc{MFillH} on both grid graphs and the PACE dataset. Except for the $5\times 5$ grid (where \textsc{ReFill} matches both heuristics), \textsc{ReFill} outperforms them on all grid sizes by up to $13.6\%$ (vs.\ \textsc{MDH}) and $9.8\%$ (vs.\ \textsc{MFillH}). 

On the PACE dataset, \textsc{ReFill} generally achieves better fill-in costs than \textsc{MDH} (up to $18.6\%$ improvement) and \textsc{MFillH} (up to $5.4\%$), except on $\textsc{92.graph}$, where \textsc{ReFill} underperforms \textsc{MFillH} by $1.5\%$. One of the reasons that \textsc{ReFill} underperforms on \textsc{92.GRAPH} was that during elimination, the number of vertices with minimum degree or minimum fill-in were comparatively high, which leads to the same action space explosion without masking. This often caused the agent to get stuck at a local optima.  
\subsection{Generalization Experiment}
\cref{genexp} discusses the generalization experiment in more details. On average,  \textsc{ReFill} learns elimination heuristics that give an elimination order that is a 2.21\% improvement over $\textsc{MDH}$ and 1.05\% improvement over \textsc{MFillH}.

\subsection{Ablation Study on Effect of Masking}
\begin{figure}[h]
    \centering
    \includegraphics[width=0.6\linewidth]{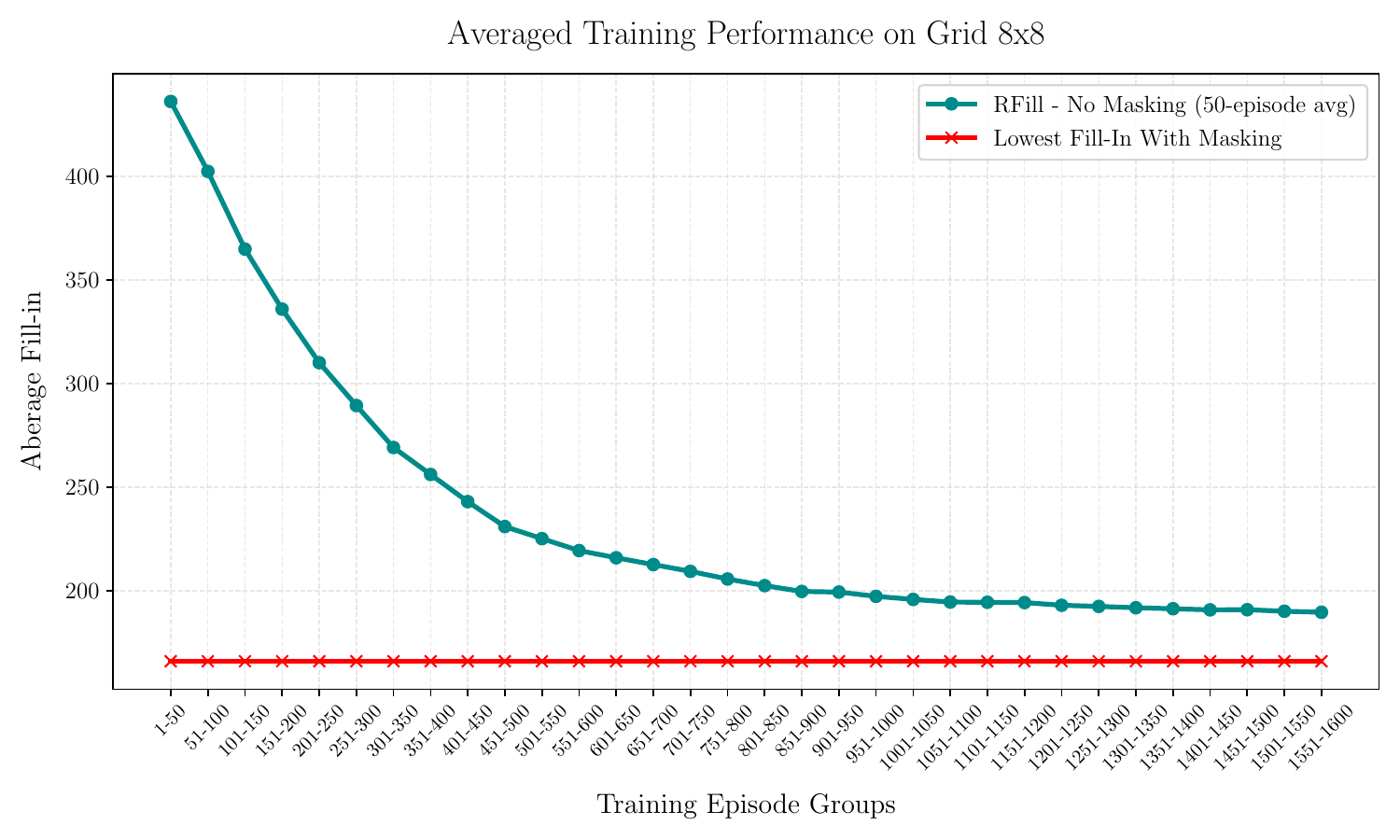}
    \caption{Average fill-in (lower is better) on $8\times 8$ grid, comparing masking vs.\ no masking during training.}
    \label{Gridn8ablation}
\end{figure}

\cref{Gridn8ablation} shows the effect of masking vs not-masking the action space on the fill-in reached by for the graph \textsc{Grid} $8\times 8$. For an ablation study on the effect of masking on training, see \cref{ablationstudysection}. As seen, not only does masking help with significantly faster convergence to better fill-in than both heuristics, not masking can also cause the agent to get stuck in a local-minima elimination ordering that is far from the optimal solution. 

\subsection{Runtime Overhead}
\textsc{ReFill} has a one-time training cost but infers node orderings efficiently once trained. In contrast, \textsc{MDH} and \textsc{MFillH} require negligible setup but can be less accurate. For repeated solves on similar graphs, the amortized cost of \textsc{ReFill} becomes favorable. Throughout, training took less than 30 minutes, with most graphs training in 15 minutes.

As for the generalization experiment, after training, the inference speed of all heuristics were  comparable, which further reinforces that GNN-based elimination orders can hopefully be deployed in practical solvers in the future. 

\subsection{Summary}
Overall, \textsc{ReFill} demonstrates that a tailored RL approach with GCN-based policies and action masking can outperform standard heuristics on both grid graphs and real-world PACE instances. Masking notably improves convergence and solution quality, while moderate GCN depth and feature dimension suffice to capture the local graph information crucial for effective elimination orders. Once trained, \textsc{ReFill} generates improved orderings with minimal inference cost, and can generalize on new unseen graph instances drawn from the same graph distribution as the training data.

\section{Future Work and Limitations}

While our results demonstrate the promise of learning-based methods for fill-in minimization, several challenges remain. First, deploying these methods in practice, particularly in single-shot settings where a solver must handle arbitrary matrices without prior training, requires a lightweight, generalizable GCN model with a \textit{single set} of parameters. However, training such a model across diverse graphs proved unstable, likely due to structural variations between datasets that hinder consistent learning. However, as we showed, when the graphs are sampled from the \textit{same distribution} (say $G(n, p)$), then it is possible to learn a single set of parameters that outperforms both heuristics on average. 
Future research could explore transfer learning or domain adaptation techniques to stabilize training across graph distributions. Second, while masking actions to prioritize minimum degree or fill-in candidates accelerates convergence, it risks excluding elimination orders that diverge from these heuristics. Yet, removing masking often traps the agent in local minima as shown in our ablation studies. Developing adaptive masking strategies that balance exploration and exploitation, for instance, by gradually relaxing masking constraints as training progresses, could mitigate this trade-off. Finally, scalability remains a critical barrier. Extending this work to million-scale systems will require more scalable GNN architectures and integration with distributed optimizers and solvers. Addressing these challenges will be essential to bridge the gap between theoretical advances and real-world deployment.

\bibliographystyle{plain}

\bibliography{section_reference}

\begin{thebibliography}{10}

\bibitem{DBLP:journals/corr/Altae-TranRPP16}
Han Altae{-}Tran, Bharath Ramsundar, Aneesh~S. Pappu, and Vijay~S. Pande.
\newblock Low data drug discovery with one-shot learning.
\newblock {\em CoRR}, abs/1611.03199, 2016.

\bibitem{tiebreaking}
Patrick~R. Amestoy, Timothy~A. Davis, and Iain~S. Duff.
\newblock An approximate minimum degree ordering algorithm.
\newblock {\em SIAM Journal on Matrix Analysis and Applications}, 17(4):886--905, 1996.

\bibitem{DBLP:journals/corr/abs-2306-16976}
Ahmed Begga, Francisco Escolano, Miguel~Angel Lozano, and Edwin~R. Hancock.
\newblock Diffusion-jump gnns: Homophiliation via learnable metric filters.
\newblock {\em CoRR}, abs/2306.16976, 2023.

\bibitem{bengio2021machine}
Yoshua Bengio, Andrea Lodi, and Antoine Prouvost.
\newblock Machine learning for combinatorial optimization: A methodological tour d’horizon.
\newblock {\em European Journal of Operational Research}, 290(2):405--421, 2021.

\bibitem{bliznets2020lower}
Ivan Bliznets, Marek Cygan, Pawe{\l} Komosa, Micha{\l} Pilipczuk, and Luk{\'a}{\v{s}} Mach.
\newblock Lower bounds for the parameterized complexity of minimum fill-in and other completion problems.
\newblock {\em ACM Transactions on Algorithms (TALG)}, 16(2):1--31, 2020.

\bibitem{matrixmarket}
Ronald~F. Boisvert, Roldan Pozo, Karin Remington, Richard~F. Barrett, and Jack~J. Dongarra.
\newblock Matrix market: a web resource for test matrix collections.
\newblock In {\em Proceedings of the IFIP TC2/WG2.5 Working Conference on Quality of Numerical Software: Assessment and Enhancement}, page 125–137, GBR, 1997. Chapman \& Hall, Ltd.

\bibitem{bollhofer2020state}
Matthias Bollh{\"o}fer, Olaf Schenk, Radim Janalik, Steve Hamm, and Kiran Gullapalli.
\newblock State-of-the-art sparse direct solvers.
\newblock {\em Parallel algorithms in computational science and engineering}, pages 3--33, 2020.

\bibitem{brezinski2022journey}
Claude Brezinski, G{\'e}rard Meurant, and Michela Redivo-Zaglia.
\newblock {\em A Journey through the History of Numerical Linear Algebra}.
\newblock SIAM, 2022.

\bibitem{inapprox}
Yixin Cao and R.B. Sandeep.
\newblock Minimum fill-in: Inapproximability and almost tight lower bounds.
\newblock {\em Information and Computation}, 271:104514, 2020.

\bibitem{DBLP:conf/icml/ChenWHDL20}
Ming Chen, Zhewei Wei, Zengfeng Huang, Bolin Ding, and Yaliang Li.
\newblock Simple and deep graph convolutional networks.
\newblock In {\em Proceedings of the 37th International Conference on Machine Learning, {ICML} 2020, 13-18 July 2020, Virtual Event}, volume 119 of {\em Proceedings of Machine Learning Research}, pages 1725--1735. {PMLR}, 2020.

\bibitem{DBLP:conf/iclr/ChenLB19}
Zhengdao Chen, Lisha Li, and Joan Bruna.
\newblock Supervised community detection with line graph neural networks.
\newblock In {\em 7th International Conference on Learning Representations, {ICLR} 2019, New Orleans, LA, USA, May 6-9, 2019}. OpenReview.net, 2019.

\bibitem{crespelle2023survey}
Christophe Crespelle, P{\aa}l~Gr{\o}n{\aa}s Drange, Fedor~V Fomin, and Petr Golovach.
\newblock A survey of parameterized algorithms and the complexity of edge modification.
\newblock {\em Computer Science Review}, 48:100556, 2023.

\bibitem{fastmindegree}
Robert Cummings, Matthew Fahrbach, and Animesh Fatehpuria.
\newblock A fast minimum degree algorithm and matching lower bound.
\newblock In {\em Proceedings of the Thirty-Second Annual ACM-SIAM Symposium on Discrete Algorithms}, SODA '21, page 724–734, USA, 2021. Society for Industrial and Applied Mathematics.

\bibitem{victorgnn}
Victor-Alexandru Darvariu, Stephen Hailes, and Mirco Musolesi.
\newblock Graph reinforcement learning for combinatorial optimization: A survey and unifying perspective, 2024.

\bibitem{pace}
Holger Dell, Christian Komusiewicz, Nimrod Talmon, and Mathias Weller.
\newblock {The PACE 2017 Parameterized Algorithms and Computational Experiments Challenge: The Second Iteration}.
\newblock In Daniel Lokshtanov and Naomi Nishimura, editors, {\em 12th International Symposium on Parameterized and Exact Computation (IPEC 2017)}, volume~89 of {\em Leibniz International Proceedings in Informatics (LIPIcs)}, pages 30:1--30:12, Dagstuhl, Germany, 2018. Schloss Dagstuhl -- Leibniz-Zentrum f{\"u}r Informatik.

\bibitem{DBLP:journals/corr/abs-2003-00982}
Vijay~Prakash Dwivedi, Chaitanya~K. Joshi, Thomas Laurent, Yoshua Bengio, and Xavier Bresson.
\newblock Benchmarking graph neural networks.
\newblock {\em CoRR}, abs/2003.00982, 2020.

\bibitem{georgenesteddissection}
Alan George.
\newblock Nested dissection of a regular finite element mesh.
\newblock {\em SIAM Journal on Numerical Analysis}, 10(2):345--363, 1973.

\bibitem{DBLP:conf/iclr/HuLGZLPL20}
Weihua Hu, Bowen Liu, Joseph Gomes, Marinka Zitnik, Percy Liang, Vijay~S. Pande, and Jure Leskovec.
\newblock Strategies for pre-training graph neural networks.
\newblock In {\em 8th International Conference on Learning Representations, {ICLR} 2020, Addis Ababa, Ethiopia, April 26-30, 2020}. OpenReview.net, 2020.

\bibitem{maskedppo}
Shengyi Huang and Santiago Ontañón.
\newblock A closer look at invalid action masking in policy gradient algorithms.
\newblock 2020.

\bibitem{DBLP:conf/bigdataconf/IzadiFSL20}
Mohammad~Rasool Izadi, Yihao Fang, Robert Stevenson, and Lizhen Lin.
\newblock Optimization of graph neural networks with natural gradient descent.
\newblock In Xintao Wu, Chris Jermaine, Li~Xiong, Xiaohua Hu, Olivera Kotevska, Siyuan Lu, Weija Xu, Srinivas Aluru, Chengxiang Zhai, Eyhab Al{-}Masri, Zhiyuan Chen, and Jeff Saltz, editors, {\em 2020 {IEEE} International Conference on Big Data {(IEEE} BigData 2020), Atlanta, GA, USA, December 10-13, 2020}, pages 171--179. {IEEE}, 2020.

\bibitem{joshi2019efficient}
Chaitanya~K Joshi, Thomas Laurent, and Xavier Bresson.
\newblock An efficient graph convolutional network technique for the travelling salesman problem.
\newblock {\em arXiv preprint arXiv:1906.01227}, 2019.

\bibitem{khalil2017learning}
Elias Khalil, Hanjun Dai, Yuyu Zhang, Bistra Dilkina, and Le~Song.
\newblock Learning combinatorial optimization algorithms over graphs.
\newblock {\em Advances in Neural Information Processing Systems (NeurIPS)}, 30, 2017.

\bibitem{DBLP:journals/corr/KipfW16a}
Thomas~N. Kipf and Max Welling.
\newblock Variational graph auto-encoders.
\newblock {\em CoRR}, abs/1611.07308, 2016.

\bibitem{DBLP:conf/iclr/KipfW17}
Thomas~N. Kipf and Max Welling.
\newblock Semi-supervised classification with graph convolutional networks.
\newblock In {\em 5th International Conference on Learning Representations, {ICLR} 2017, Toulon, France, April 24-26, 2017, Conference Track Proceedings}. OpenReview.net, 2017.

\bibitem{kool2019attention}
Wouter Kool, Herke van Hoof, and Max Welling.
\newblock Attention, learn to solve routing problems!
\newblock {\em arXiv preprint arXiv:1803.08475}, 2019.

\bibitem{DBLP:journals/tsp/LevieMBB19}
Ron Levie, Federico Monti, Xavier Bresson, and Michael~M. Bronstein.
\newblock Cayleynets: Graph convolutional neural networks with complex rational spectral filters.
\newblock {\em {IEEE} Trans. Signal Process.}, 67(1):97--109, 2019.

\bibitem{DBLP:journals/corr/LiTBZ15}
Yujia Li, Daniel Tarlow, Marc Brockschmidt, and Richard~S. Zemel.
\newblock Gated graph sequence neural networks.
\newblock In Yoshua Bengio and Yann LeCun, editors, {\em 4th International Conference on Learning Representations, {ICLR} 2016, San Juan, Puerto Rico, May 2-4, 2016, Conference Track Proceedings}, 2016.

\bibitem{multiplemdh}
Joseph W.~H. Liu.
\newblock Modification of the minimum-degree algorithm by multiple elimination.
\newblock {\em ACM Trans. Math. Softw.}, 11(2):141–153, June 1985.

\bibitem{DBLP:journals/corr/abs-2109-05641}
Sitao Luan, Chenqing Hua, Qincheng Lu, Jiaqi Zhu, Mingde Zhao, Shuyuan Zhang, Xiao{-}Wen Chang, and Doina Precup.
\newblock Is heterophily {A} real nightmare for graph neural networks to do node classification?
\newblock {\em CoRR}, abs/2109.05641, 2021.

\bibitem{DBLP:conf/sigir/0001GRTY20}
Yao Ma, Ziyi Guo, Zhaochun Ren, Jiliang Tang, and Dawei Yin.
\newblock Streaming graph neural networks.
\newblock In Jimmy~X. Huang, Yi~Chang, Xueqi Cheng, Jaap Kamps, Vanessa Murdock, Ji{-}Rong Wen, and Yiqun Liu, editors, {\em Proceedings of the 43rd International {ACM} {SIGIR} conference on research and development in Information Retrieval, {SIGIR} 2020, Virtual Event, China, July 25-30, 2020}, pages 719--728. {ACM}, 2020.

\bibitem{markowitz}
Harry~M. Markowitz.
\newblock The elimination form of the inverse and its application to linear programming.
\newblock {\em Management Science}, 3(3):255--269, 1957.

\bibitem{mazyavkina2021reinforcement}
Nina Mazyavkina, Sergei Sviridov, Sergei Ivanov, and Evgeny Burnaev.
\newblock Reinforcement learning for combinatorial optimization: A survey.
\newblock {\em Computers \& Operations Research}, 134:105400, 2021.

\bibitem{extrmalgraphalpha}
Abbas Mehrabian, Ankit Anand, Hyunjik Kim, Nicolas Sonnerat, Matej Balog, Gheorghe Comanici, Tudor Berariu, Andrew Lee, Anian Ruoss, Anna Bulanova, Daniel Toyama, Sam Blackwell, Bernardino~Romera Paredes, Petar Veličković, Laurent Orseau, Joonkyung Lee, Anurag~Murty Naredla, Doina Precup, and Adam~Zsolt Wagner.
\newblock Finding increasingly large extremal graphs with alphazero and tabu search, 2023.

\bibitem{mnih2015human}
Volodymyr Mnih, Koray Kavukcuoglu, David Silver, Andrei~A. Rusu, Joel Veness, Marc~G. Bellemare, Alex Graves, Martin Riedmiller, Andreas~K. Fidjeland, Georg Ostrovski, Stig Petersen, Charles Beattie, Amir Sadik, Ioannis Antonoglou, Helen King, Dharshan Kumaran, Daan Wierstra, Shane Legg, and Demis Hassabis.
\newblock Human-level control through deep reinforcement learning.
\newblock In {\em Nature}, volume 518, pages 529--533, 2015.

\bibitem{Mucha2006}
Marcin Mucha and Piotr Sankowski.
\newblock Maximum matchings in planar graphs via gaussian elimination.
\newblock {\em Algorithmica}, 45(1):3--20, May 2006.

\bibitem{DBLP:journals/corr/abs-2302-04181}
Luis M{\"{u}}ller, Mikhail Galkin, Christopher Morris, and Ladislav Ramp{\'{a}}sek.
\newblock Attending to graph transformers.
\newblock {\em CoRR}, abs/2302.04181, 2023.

\bibitem{DBLP:conf/nips/NicolicioiuDL19}
Andrei~Liviu Nicolicioiu, Iulia Duta, and Marius Leordeanu.
\newblock Recurrent space-time graph neural networks.
\newblock In Hanna~M. Wallach, Hugo Larochelle, Alina Beygelzimer, Florence d'Alch{\'{e}}{-}Buc, Emily~B. Fox, and Roman Garnett, editors, {\em Advances in Neural Information Processing Systems 32: Annual Conference on Neural Information Processing Systems 2019, NeurIPS 2019, December 8-14, 2019, Vancouver, BC, Canada}, pages 12818--12830, 2019.

\bibitem{stable-baselines3}
Antonin Raffin, Ashley Hill, Adam Gleave, Anssi Kanervisto, Maximilian Ernestus, and Noah Dormann.
\newblock Stable-baselines3: Reliable reinforcement learning implementations.
\newblock {\em Journal of Machine Learning Research}, 22(268):1--8, 2021.

\bibitem{DBLP:conf/nips/RongBXX0HH20}
Yu~Rong, Yatao Bian, Tingyang Xu, Weiyang Xie, Ying Wei, Wenbing Huang, and Junzhou Huang.
\newblock Self-supervised graph transformer on large-scale molecular data.
\newblock In Hugo Larochelle, Marc'Aurelio Ranzato, Raia Hadsell, Maria{-}Florina Balcan, and Hsuan{-}Tien Lin, editors, {\em Advances in Neural Information Processing Systems 33: Annual Conference on Neural Information Processing Systems 2020, NeurIPS 2020, December 6-12, 2020, virtual}, 2020.

\bibitem{ROSE1970597}
Donald~J Rose.
\newblock Triangulated graphs and the elimination process.
\newblock {\em Journal of Mathematical Analysis and Applications}, 32(3):597--609, 1970.

\bibitem{ROSE1972183}
Donald~J. Rose.
\newblock A graph-theoretic study of the numerical solution of sparse positive definite systems of linear equations.
\newblock In RONALD~C. READ, editor, {\em Graph Theory and Computing}, pages 183--217. Academic Press, 1972.

\bibitem{rosetarjan}
Donald~J. Rose, R.~Endre Tarjan, and George~S. Lueker.
\newblock Algorithmic aspects of vertex elimination on graphs.
\newblock {\em SIAM Journal on Computing}, 5(2):266--283, 1976.

\bibitem{networkrepo}
Ryan~A. Rossi and Nesreen~K. Ahmed.
\newblock Networkrepository: An interactive data repository with multi-scale visual analytics, 2014.

\bibitem{scarselli2008graph}
Franco Scarselli, Marco Gori, Ah~Chung Tsoi, Markus Hagenbuchner, and Gabriele Monfardini.
\newblock The graph neural network model.
\newblock {\em IEEE transactions on neural networks}, 20(1):61--80, 2008.

\bibitem{schulman2017proximal}
John Schulman, Filip Wolski, Prafulla Dhariwal, Alec Radford, and Oleg Klimov.
\newblock Proximal policy optimization algorithms.
\newblock In {\em arXiv preprint arXiv:1707.06347}, 2017.

\bibitem{DBLP:conf/icml/ShirzadVVSS23}
Hamed Shirzad, Ameya Velingker, Balaji Venkatachalam, Danica~J. Sutherland, and Ali~Kemal Sinop.
\newblock Exphormer: Sparse transformers for graphs.
\newblock In Andreas Krause, Emma Brunskill, Kyunghyun Cho, Barbara Engelhardt, Sivan Sabato, and Jonathan Scarlett, editors, {\em International Conference on Machine Learning, {ICML} 2023, 23-29 July 2023, Honolulu, Hawaii, {USA}}, volume 202 of {\em Proceedings of Machine Learning Research}, pages 31613--31632. {PMLR}, 2023.

\bibitem{silver2016mastering}
David Silver, Aja Huang, Chris~J. Maddison, Arthur Guez, Laurent Sifre, George Van~Den Driessche, Julian Schrittwieser, Ioannis Antonoglou, Veda Panneershelvam, Marc Lanctot, et~al.
\newblock Mastering the game of go with deep neural networks and tree search.
\newblock {\em Nature}, 529(7587):484--489, 2016.

\bibitem{silver2017mastering}
David Silver, Julian Schrittwieser, Karen Simonyan, Ioannis Antonoglou, Aja Huang, Arthur Guez, Thomas Hubert, Lucas Baker, Matthew Lai, Adrian Bolton, et~al.
\newblock Mastering the game of go without human knowledge.
\newblock {\em Nature}, 550(7676):354--359, 2017.

\bibitem{igorgnnjscheduling}
Igor~G. Smit, Jianan Zhou, Robbert Reijnen, Yaoxin Wu, Jian Chen, Cong Zhang, Zaharah Bukhsh, Yingqian Zhang, and Wim Nuijten.
\newblock Graph neural networks for job shop scheduling problems: A survey, 2024.

\bibitem{DBLP:journals/air/SunLLZMRW23}
Chengcheng Sun, Chenhao Li, Xiang Lin, Tianji Zheng, Fanrong Meng, Xiaobin Rui, and Zhixiao Wang.
\newblock Attention-based graph neural networks: a survey.
\newblock {\em Artif. Intell. Rev.}, 56({S2}):2263--2310, 2023.

\bibitem{sutton2018reinforcement}
Richard~S. Sutton and Andrew~G. Barto.
\newblock {\em Reinforcement Learning: An Introduction}.
\newblock MIT Press, 2nd edition, 2018.

\bibitem{tang2021or}
Yunhao Tang, Shipra Agrawal, and Yuri Faenza.
\newblock Or-gym: A reinforcement learning library for operations research problems.
\newblock In {\em NeurIPS Datasets and Benchmarks}, 2021.

\bibitem{DBLP:journals/corr/abs-1710-10903}
Petar Velickovic, Guillem Cucurull, Arantxa Casanova, Adriana Romero, Pietro Li{\`{o}}, and Yoshua Bengio.
\newblock Graph attention networks.
\newblock {\em CoRR}, abs/1710.10903, 2017.

\bibitem{DBLP:journals/tnn/WuPCLZY21}
Zonghan Wu, Shirui Pan, Fengwen Chen, Guodong Long, Chengqi Zhang, and Philip~S. Yu.
\newblock A comprehensive survey on graph neural networks.
\newblock {\em {IEEE} Trans. Neural Networks Learn. Syst.}, 32(1):4--24, 2021.

\bibitem{npcomplete}
Mihalis Yannakakis.
\newblock Computing the minimum fill-in is np-complete.
\newblock {\em SIAM Journal on Algebraic Discrete Methods}, 2(1):77--79, 1981.

\bibitem{DBLP:journals/corr/abs-1802-08773}
Jiaxuan You, Rex Ying, Xiang Ren, William~L. Hamilton, and Jure Leskovec.
\newblock Graphrnn: {A} deep generative model for graphs.
\newblock {\em CoRR}, abs/1802.08773, 2018.

\bibitem{DBLP:conf/nips/YunJKKK19}
Seongjun Yun, Minbyul Jeong, Raehyun Kim, Jaewoo Kang, and Hyunwoo~J. Kim.
\newblock Graph transformer networks.
\newblock In Hanna~M. Wallach, Hugo Larochelle, Alina Beygelzimer, Florence d'Alch{\'{e}}{-}Buc, Emily~B. Fox, and Roman Garnett, editors, {\em Advances in Neural Information Processing Systems 32: Annual Conference on Neural Information Processing Systems 2019, NeurIPS 2019, December 8-14, 2019, Vancouver, BC, Canada}, pages 11960--11970, 2019.

\bibitem{DBLP:conf/csonet/ZhangTXM18}
Si~Zhang, Hanghang Tong, Jiejun Xu, and Ross Maciejewski.
\newblock Graph convolutional networks: Algorithms, applications and open challenges.
\newblock In Xuemin Chen, Arunabha Sen, Wei~Wayne Li, and My~T. Thai, editors, {\em Computational Data and Social Networks - 7th International Conference, CSoNet 2018, Shanghai, China, December 18-20, 2018, Proceedings}, volume 11280 of {\em Lecture Notes in Computer Science}, pages 79--91. Springer, 2018.

\bibitem{DBLP:journals/tkde/ZhangCZ22}
Ziwei Zhang, Peng Cui, and Wenwu Zhu.
\newblock Deep learning on graphs: {A} survey.
\newblock {\em {IEEE} Trans. Knowl. Data Eng.}, 34(1):249--270, 2022.

\bibitem{DBLP:conf/nips/ZhuZXT21}
Zhaocheng Zhu, Zuobai Zhang, Louis{-}Pascal A.~C. Xhonneux, and Jian Tang.
\newblock Neural bellman-ford networks: {A} general graph neural network framework for link prediction.
\newblock In Marc'Aurelio Ranzato, Alina Beygelzimer, Yann~N. Dauphin, Percy Liang, and Jennifer~Wortman Vaughan, editors, {\em Advances in Neural Information Processing Systems 34: Annual Conference on Neural Information Processing Systems 2021, NeurIPS 2021, December 6-14, 2021, virtual}, pages 29476--29490, 2021.

\end{thebibliography}

\appendix
\onecolumn
\section{Implementation Details}
\label{ImplementationDetails}
The learning rate is set to $1\times 10^{-4}$ or $5\times 10^{-5}$ depending on graph size. Hyperparameters are detailed in  \cref{hyperparamsmain}; our code is in the supplemental section. Experiments are performed on a single machine with an NVIDIA A100-SXM4-80GB GPU (226\,GiB system memory). We run 5 parallel environments to collect experiences using stable baselines3~\cite{stable-baselines3} with maskable PPO~\cite{maskedppo}. The GCN’s hidden feature dimension ranges between $8$ and $32$. The network outputs a single scalar score per node, used for action selection.

\section{Hypterparameter values used.}
\label{hyperparamsmain}
Here, we describe all the commands we used to run the experiments, which include all the hyperparameters. The figures in \cref{table1} were calculated using the main.py script with --action\_masking set to 1, --policy\_sizes set to an empty list, --total\_timesteps set to 500,000, --parallel\_envs set to 5, and the following hyperparameters. 

\begin{table}[ht]
\caption{Hyperparameters used for \textsc{ReFill}.}
\label{table:hyperparams}
\vskip 0.1in
\begin{center}
\begin{small}
\begin{sc}
\begin{tabular}{lcccc}
\toprule
\textbf{Graph} & \textbf{total\_timesteps} & \textbf{parallel\_envs} & \textbf{learning\_rate} & \textbf{node\_dim} \\
\midrule
Grid 5x5    & 500000 & 5 & 0.0001 & 32 \\
Grid 6x6    & 500000 & 5 & 0.0001 & 32 \\
Grid 7x7    & 500000 & 5 & 0.0001 & 32 \\
Grid 8x8    & 500000 & 5 & 0.00005 & 32 \\
Grid 9x9    & 500000 & 5 & 0.00005 & 16 \\
Grid 10x10  & 500000 & 5 & 0.00005 & 8 \\
2.graph     & 500000 & 5 & 0.00005 & 16 \\
3.graph     & 500000 & 5 & 0.0001  & 32 \\
11.graph    & 500000 & 5 & 0.00005 & 32 \\
13.graph    & 500000 & 5 & 0.00005 & 16 \\
18.graph    & 500000 & 5 & 0.00005 & 16 \\
23.graph    & 500000 & 5 & 0.00005 & 16 \\
26.graph    & 500000 & 5 & 0.00005 & 16 \\
40.graph    & 500000 & 5 & 0.00005 & 16 \\
92.graph    & 500000 & 5 & 0.00005 & 16 \\
99.graph    & 500000 & 5 & 0.00005 & 16 \\
100.graph   & 500000 & 5 & 0.00005 & 16 \\
\bottomrule
\end{tabular}
\end{sc}
\end{small}
\end{center}
\vskip -0.1in
\end{table}

\section{Non-Masking Ablation Study}
\label{ablationstudysection}
The following ablation study for non-masking was performed using the command preceding the figure. Usually, the agent got close to the optimal fill-in, but gets stuck in a local optima without using action-masking.

\begin{lstlisting}[language=bash]
  $ python main.py datasets/grid.n8.graph --output_file non_masking_results/grid.n8.graph  
  --policy_sizes 16 16 --total_timesteps 500_000 --learning_rate 0.0001  
  --parallel_envs 5 --node_dim 8 --ent_coef 0.002 --action_masking 0
\end{lstlisting}
\begin{figure}[H]
    \centering
    \includegraphics[width=0.6\linewidth]{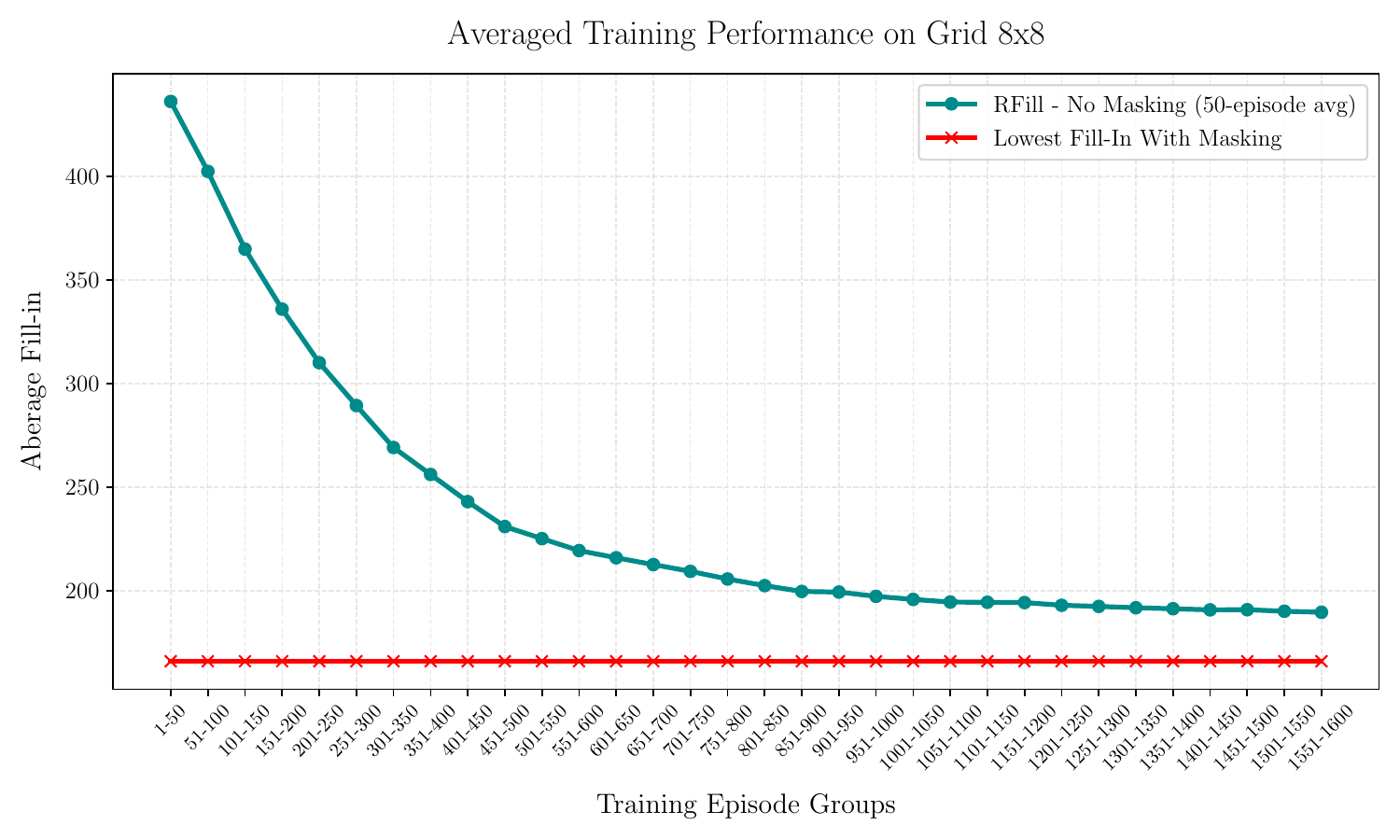}
    \caption{Average fill-in (lower is better) on $8\times 8$ grid, comparing masking vs.\ no masking during training.}
    \label{Gridn8ablation2}
\end{figure}

\begin{lstlisting}[language=bash]
  $ python main.py datasets/grid.n9.graph --output_file non_masking_results/grid.n9.graph
  --policy_sizes 32 32 --total_timesteps 500_000 --learning_rate 0.0001 
  --parallel_envs 5 --node_dim 16 --ent_coef 0.002 --action_masking 0 
\end{lstlisting}

\begin{figure}[H]
    \centering
    \includegraphics[width=0.7\linewidth]{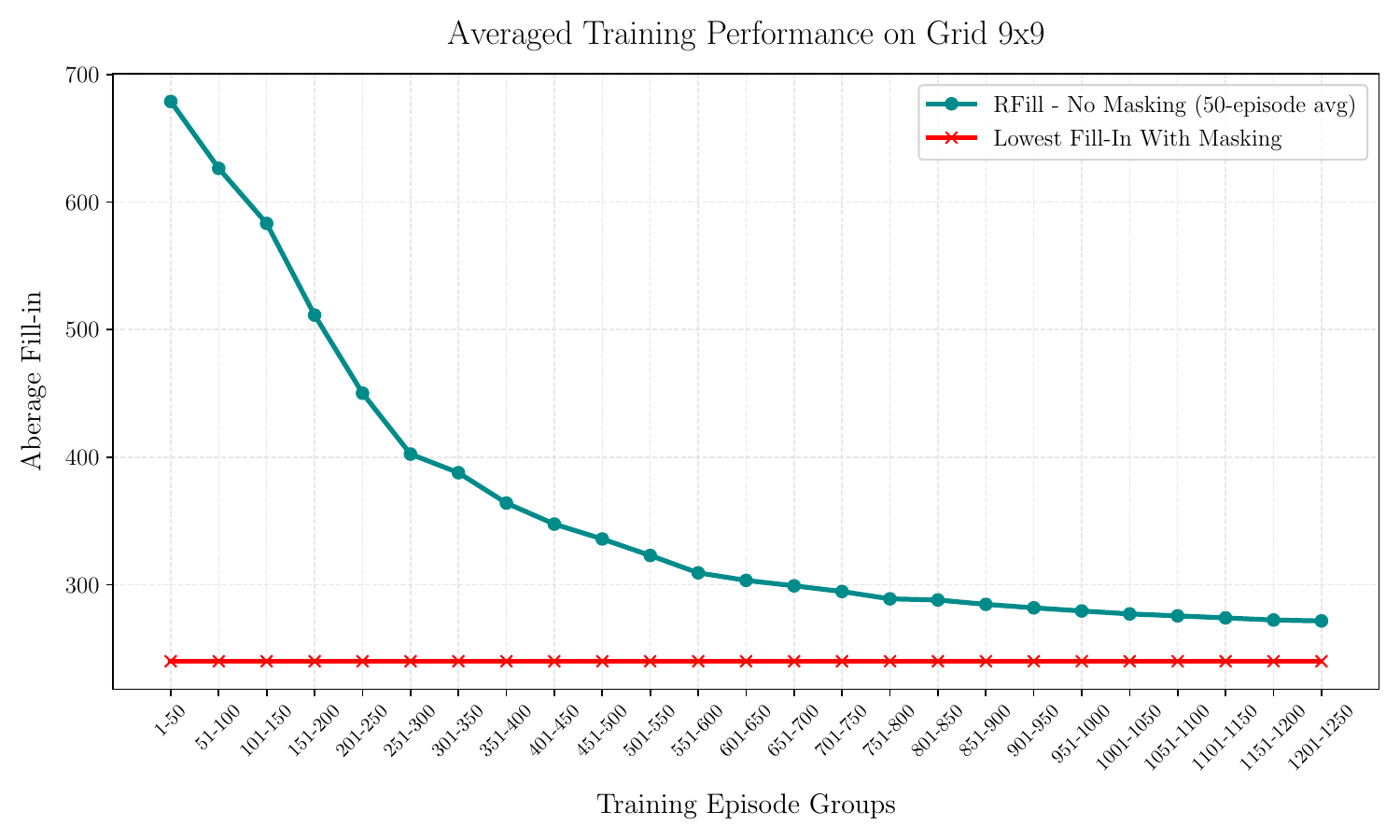}
    \caption{Average fill-in (lower is better) on $9\times 9$ grid, comparing masking vs.\ no masking during training.}
    \label{Gridn9ablation}
\end{figure}

\begin{lstlisting}[language=bash]
  $ python main.py datasets/grid.n10.graph --output_file non_masking_results/grid.n10.graph 
  --policy_sizes 32 32 --total_timesteps 500_000 --learning_rate 0.0001 
  --parallel_envs 5 --node_dim 16 --ent_coef 0.002 --action_masking 0 
\end{lstlisting}

\begin{figure}[H]
    \centering
    \includegraphics[width=0.7\linewidth]{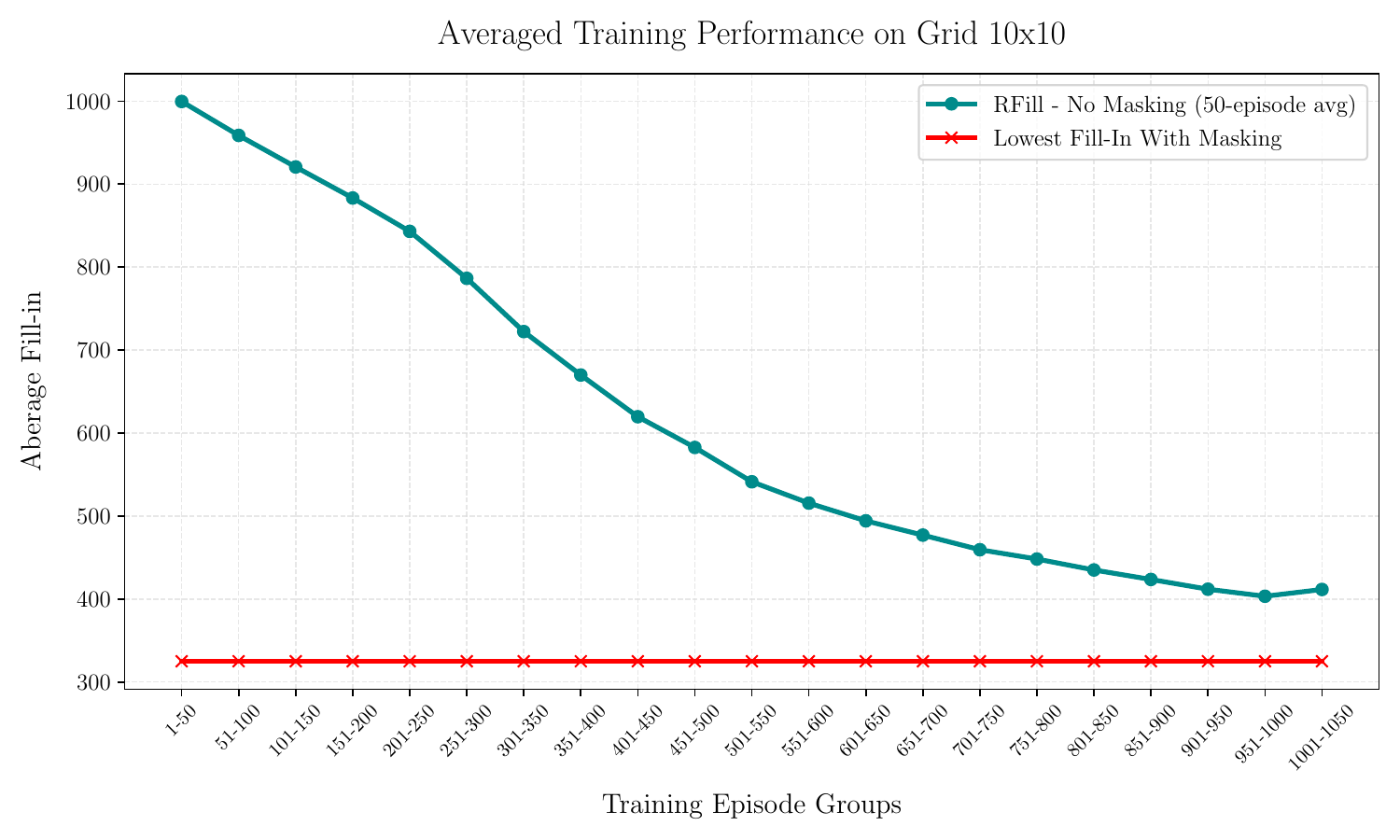}
    \caption{Average fill-in (lower is better) on $10\times 10$ grid, comparing masking vs.\ no masking during training.}
    \label{Gridn10ablation}
\end{figure}

\begin{lstlisting}[language=bash]
  $ python main.py datasets/2.graph --output_file non_masking_results/2.graph  
  --policy_sizes 16 16 --total_timesteps 500_000 --learning_rate 0.0001  
  --parallel_envs 5 --node_dim 8 --ent_coef 0.002 --action_masking 0
\end{lstlisting}

\begin{figure}[H]
    \centering
    \includegraphics[width=0.7\linewidth]{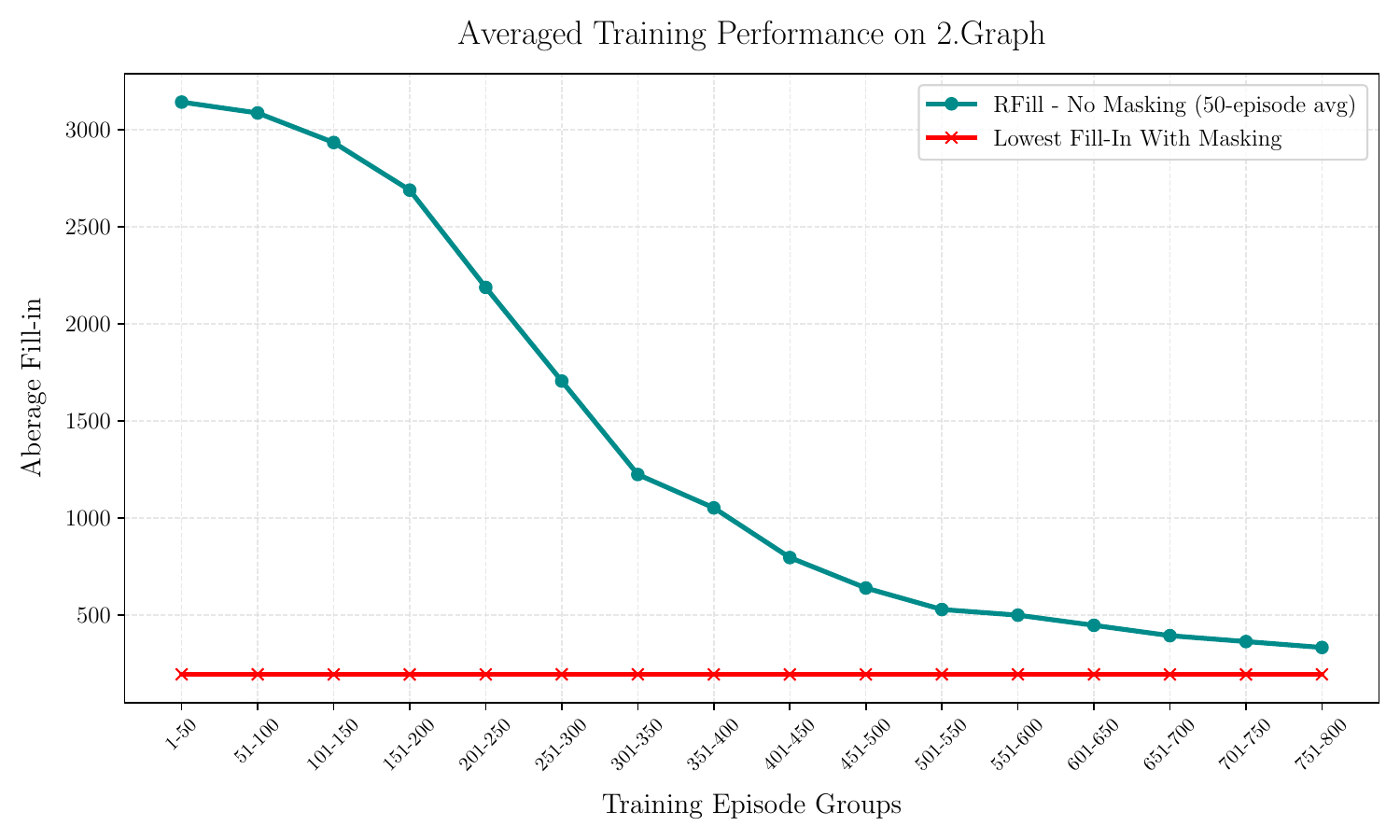}
    \caption{Average fill-in (lower is better) on $\textsc{2.graph}$, comparing masking vs.\ no masking during training.}
    \label{2Graphablation}
\end{figure}

\begin{lstlisting}[language=bash]
  $ python main.py datasets/3.graph --output_file non_masking_results/3.graph 
  --policy_sizes 32 32 --total_timesteps 500_000 --learning_rate 0.0001   
  --parallel_envs 5 --node_dim 16 --ent_coef 0.002 --action_masking 0
\end{lstlisting}

\begin{figure}[H]
    \centering
    \includegraphics[width=0.7\linewidth]{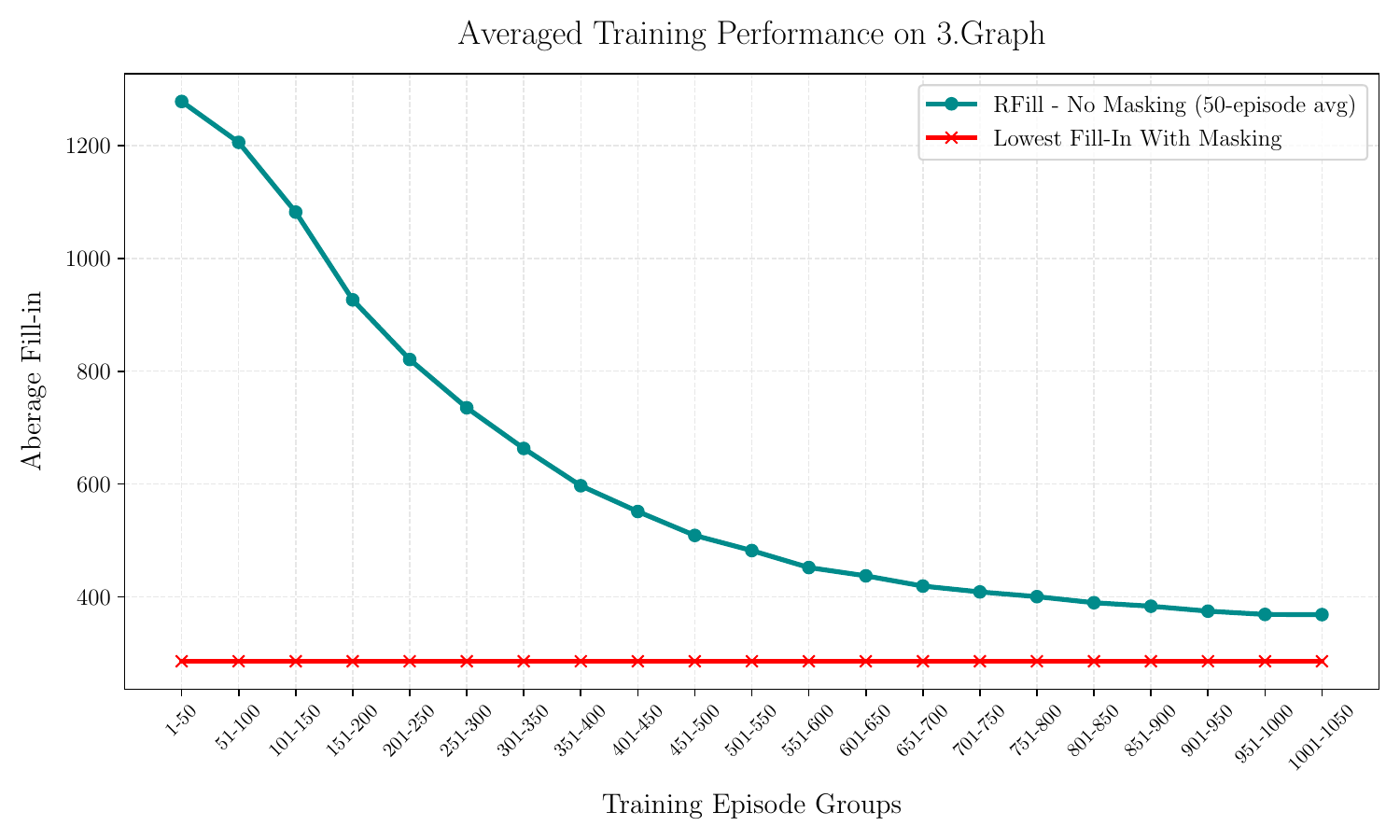}
    \caption{Average fill-in (lower is better) on $\textsc{3.graph}$, comparing masking vs.\ no masking during training.}
    \label{3Graphablation}
\end{figure}

\begin{lstlisting}[language=bash]
  $ python main.py datasets/11.graph --output_file non_masking_results/11.graph  
  --policy_sizes 32 32 --total_timesteps 500_000 --learning_rate 0.0001   
  --parallel_envs 5 --node_dim 16 --ent_coef 0.002 --action_masking 0

\end{lstlisting}

\begin{figure}[H]
    \centering
    \includegraphics[width=0.7\linewidth]{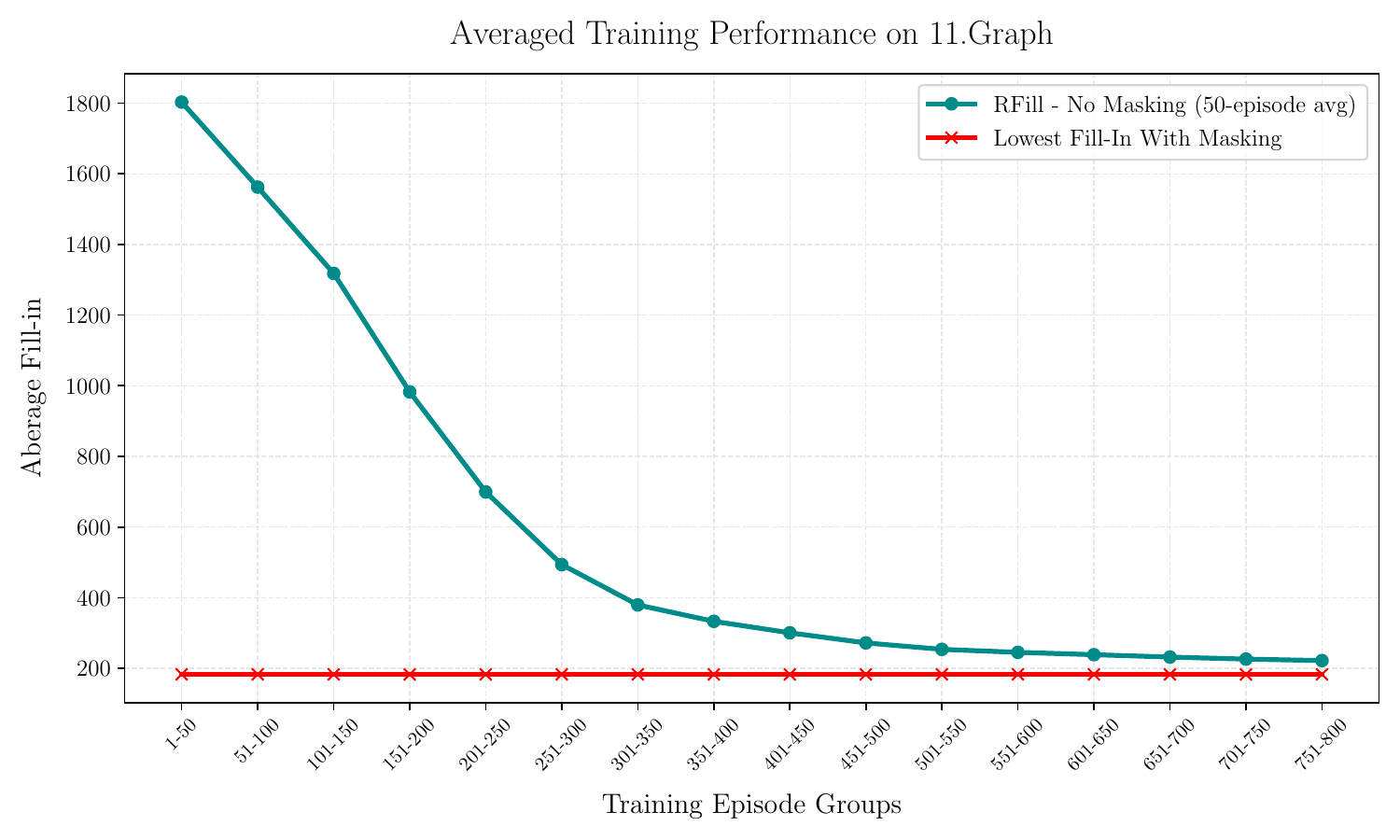}
    \caption{Average fill-in (lower is better) on $\textsc{11.graph}$, comparing masking vs.\ no masking during training.}
    \label{11Graphablation}
\end{figure}

\begin{lstlisting}[language=bash]
  $ python main.py datasets/13.graph --output_file non_masking_results/13.graph  
  --policy_sizes 32 32 --total_timesteps 500_000 --learning_rate 0.0001   
  --parallel_envs 5 --node_dim 16 --ent_coef 0.002 --action_masking 0
\end{lstlisting}

\begin{figure}[H]
    \centering
    \includegraphics[width=0.7\linewidth]{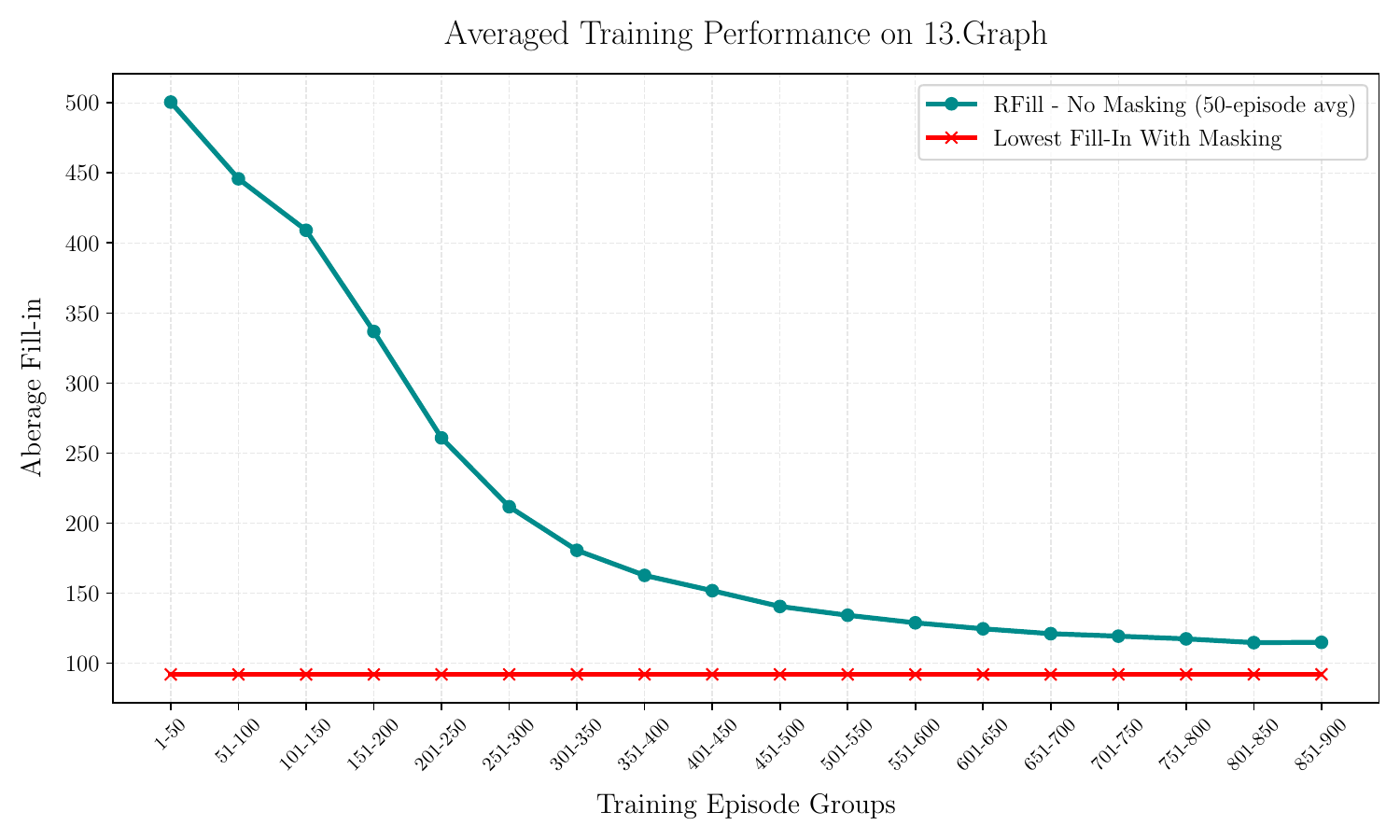}
    \caption{Average fill-in (lower is better) on $\textsc{13.graph}$, comparing masking vs.\ no masking during training.}
    \label{13Graphablation}
\end{figure}

\begin{lstlisting}[language=bash]
  $ python main.py datasets/18.graph --output_file non_masking_results/18.graph  
  --policy_sizes 32 32 --total_timesteps 500_000 --learning_rate 0.0001   
  --parallel_envs 5 --node_dim 16 --ent_coef 0.002 --action_masking 0
\end{lstlisting}

\begin{figure}[H]
    \centering
    \includegraphics[width=0.7\linewidth]{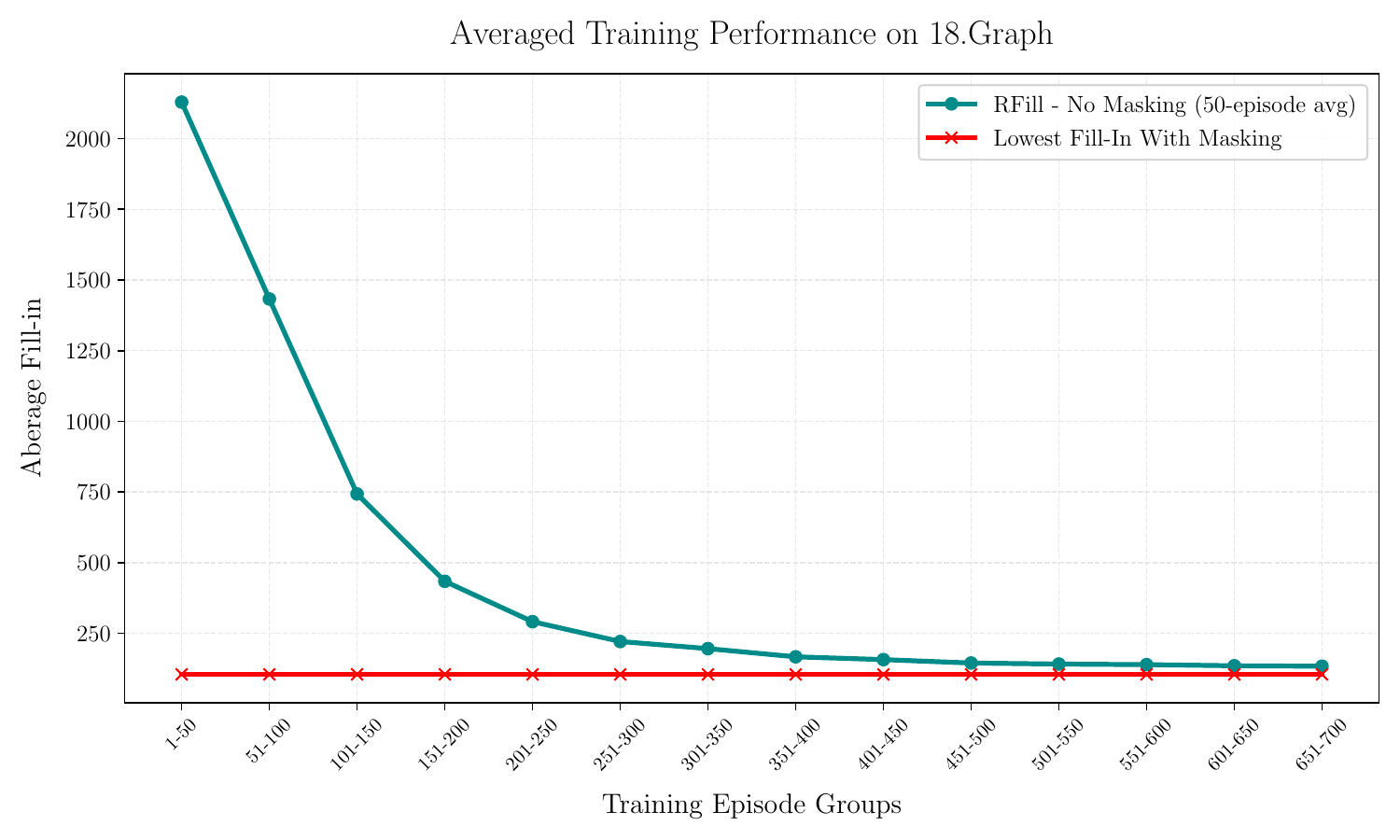}
    \caption{Average fill-in (lower is better) on $\textsc{18.graph}$, comparing masking vs.\ no masking during training.}
    \label{18Graphablation}
\end{figure}

\begin{lstlisting}[language=bash]
  $ python main.py datasets/23.graph --output_file non_masking_results/23.graph  
  --policy_sizes 64 32 --total_timesteps 1_000_000 --learning_rate 0.0001   
  --parallel_envs 10 --node_dim 16 --ent_coef 0.002 --action_masking 0
\end{lstlisting}

\begin{figure}[H]
    \centering
    \includegraphics[width=0.7\linewidth]{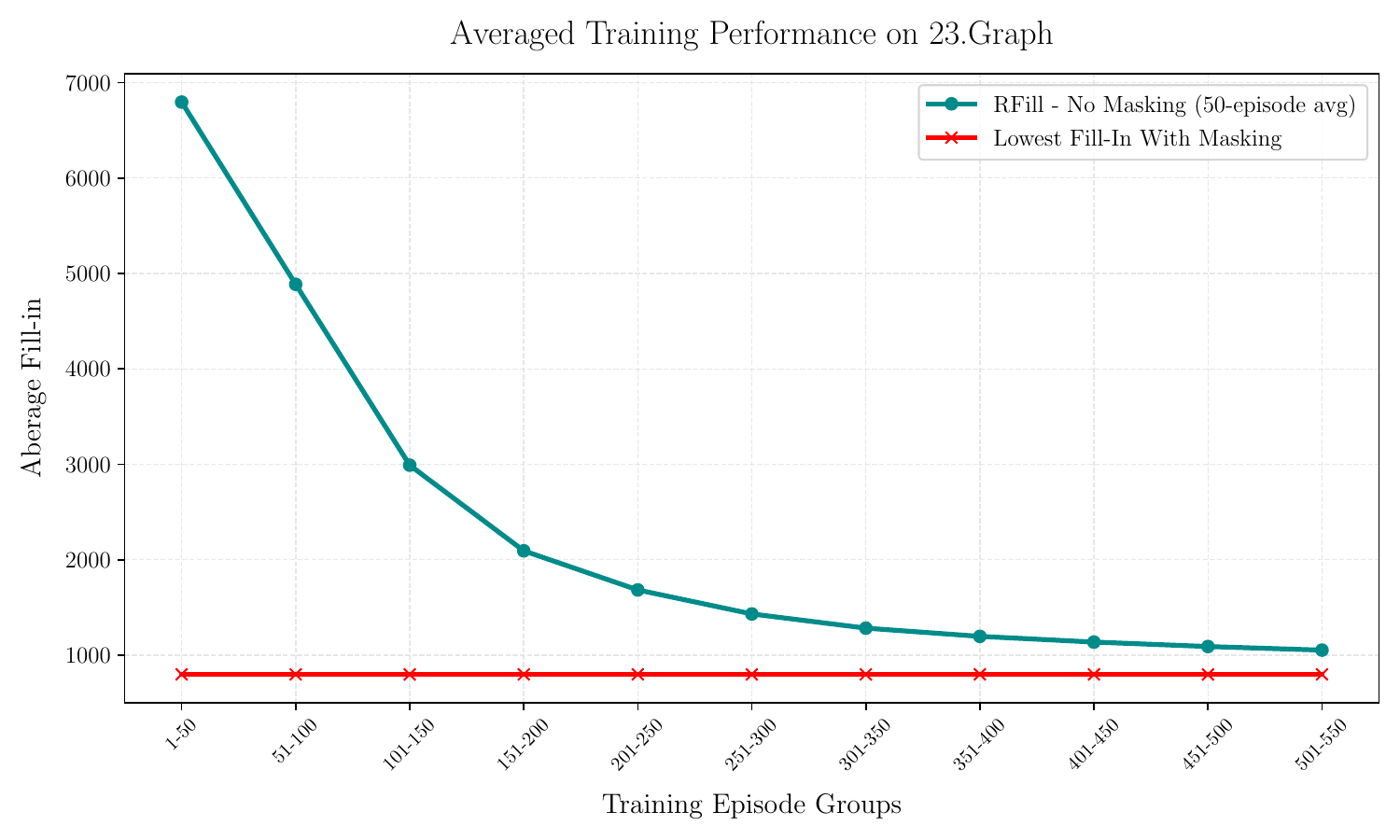}
    \caption{Average fill-in (lower is better) on $\textsc{23.graph}$, comparing masking vs.\ no masking during training.}
    \label{23Graphablation}
\end{figure}

\begin{lstlisting}[language=bash]
  $ python main.py datasets/26.graph --output_file non_masking_results/26.graph  
  --policy_sizes 32 32 --total_timesteps 500_000 --learning_rate 0.0001   
  --parallel_envs 5 --node_dim 16 --ent_coef 0.002 --action_masking 0
\end{lstlisting}

\begin{figure}[H]
    \centering
    \includegraphics[width=0.7\linewidth]{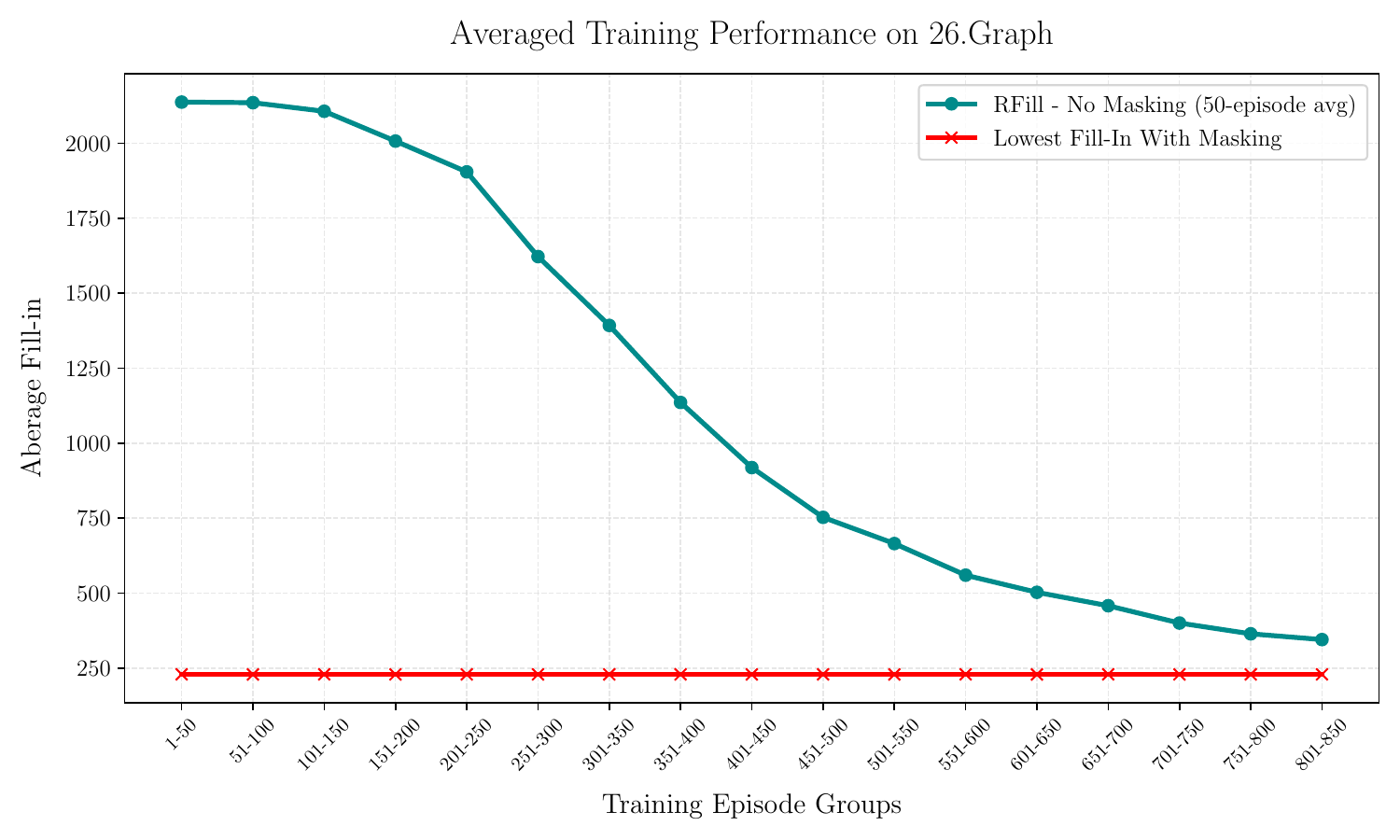}
    \caption{Average fill-in (lower is better) on $\textsc{26.graph}$, comparing masking vs.\ no masking during training.}
    \label{26Graphablation}
\end{figure}

\begin{lstlisting}[language=bash]
  $ python main.py datasets/40.graph --output_file non_masking_results/40.graph  
  --policy_sizes 64 32 --total_timesteps 1_000_000 --learning_rate 0.0001   
  --parallel_envs 10 --node_dim 16 --ent_coef 0.002 --action_masking 0
\end{lstlisting}

\begin{figure}[H]
    \centering
    \includegraphics[width=0.7\linewidth]{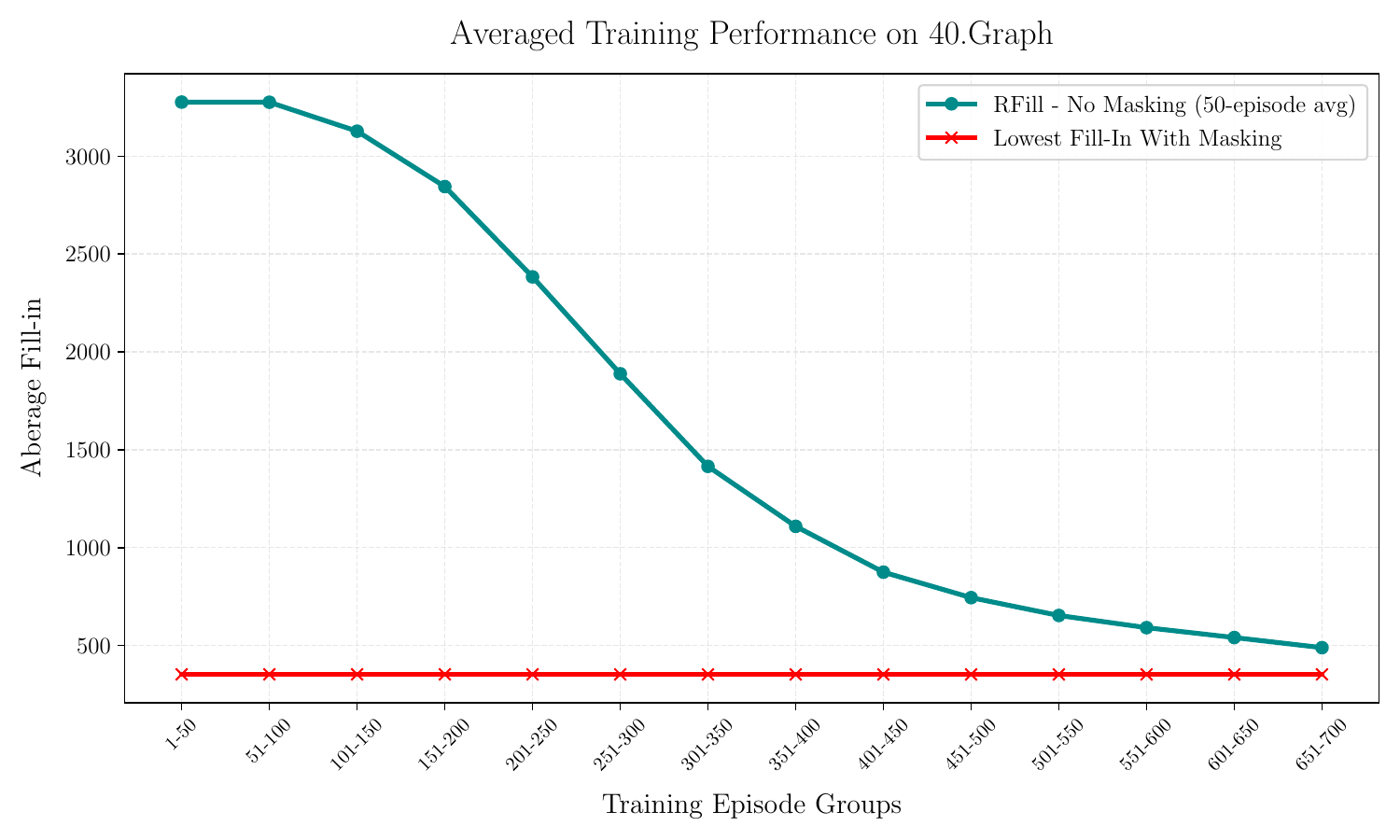}
    \caption{Average fill-in (lower is better) on $\textsc{40.graph}$, comparing masking vs.\ no masking during training.}
    \label{40Graphablation}
\end{figure}

\begin{lstlisting}[language=bash]
  $ python main.py datasets/92.graph --output_file non_masking_results/92.graph  
  --policy_sizes 64 32 --total_timesteps 1_000_000 --learning_rate 0.0001   
  --parallel_envs 10 --node_dim 16 --ent_coef 0.002 --action_masking 0
\end{lstlisting}

\begin{figure}[H]
    \centering
    \includegraphics[width=0.7\linewidth]{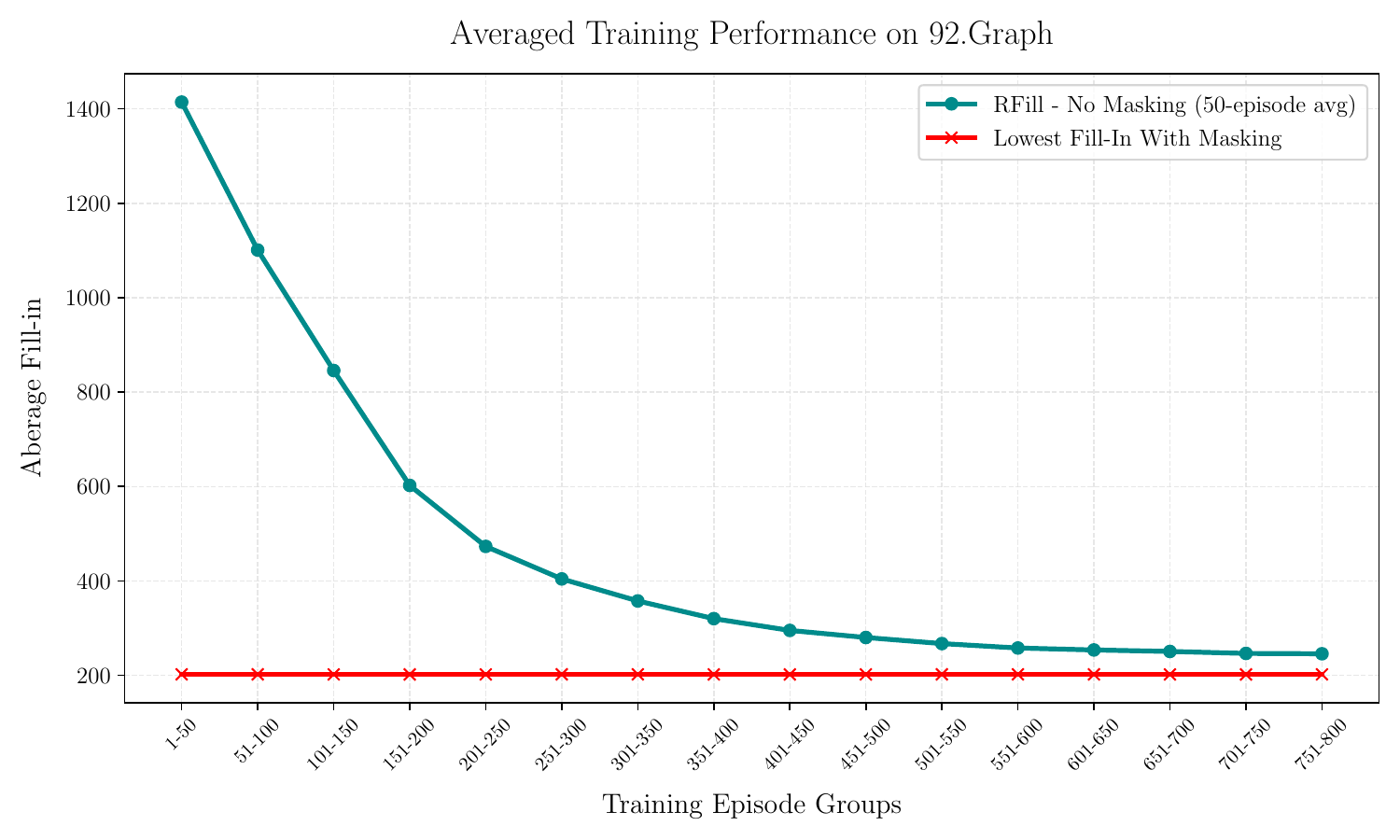}
    \caption{Average fill-in (lower is better) on $\textsc{92.graph}$, comparing masking vs.\ no masking during training.}
    \label{92Graphablation}
\end{figure}

\begin{lstlisting}[language=bash]
  $ python main.py datasets/99.graph --output_file non_masking_results/99.graph  
  --policy_sizes 64 32 --total_timesteps 1_000_000 --learning_rate 0.0001   
  --parallel_envs 10 --node_dim 16 --ent_coef 0.002 --action_masking 0
\end{lstlisting}

\begin{figure}[H]
    \centering
    \includegraphics[width=0.7\linewidth]{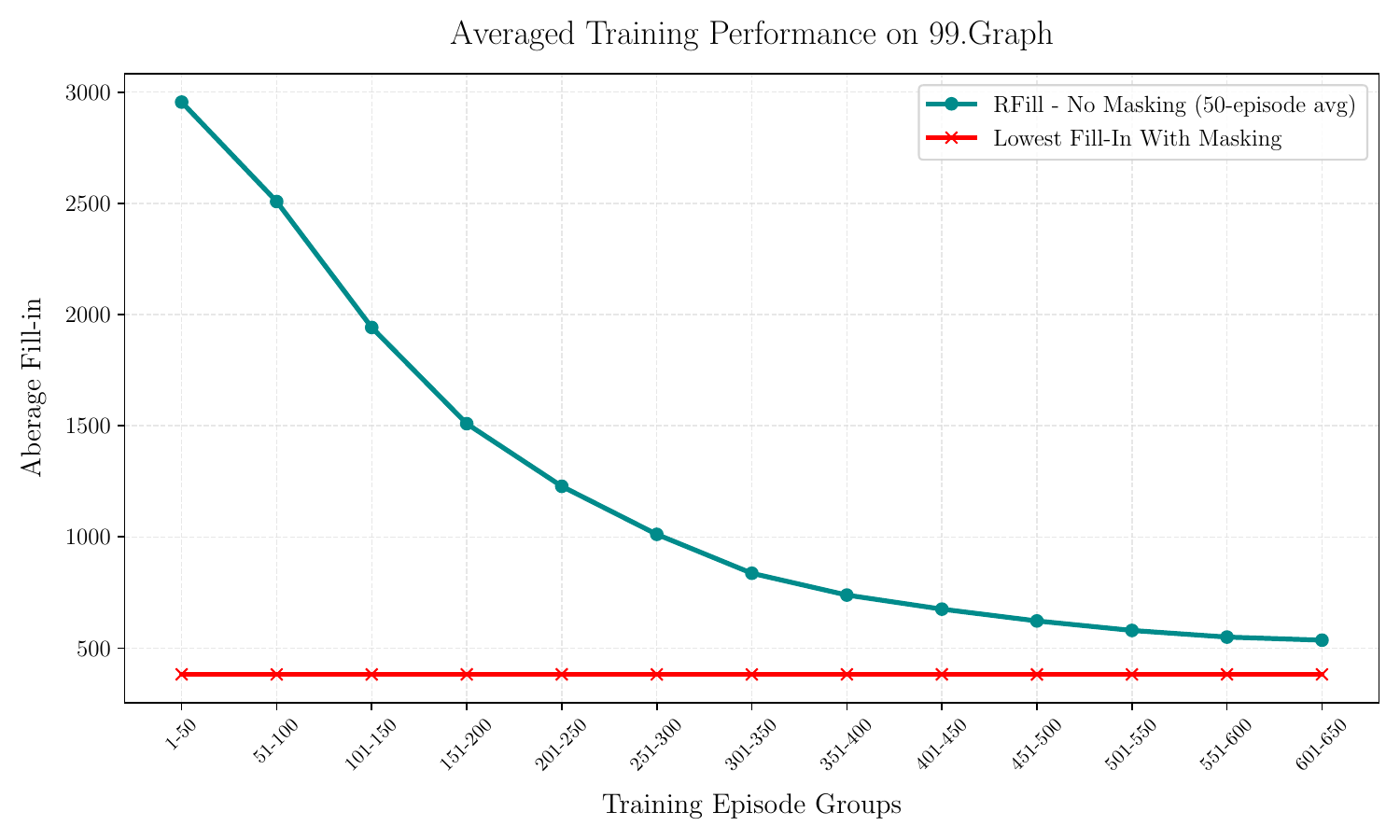}
    \caption{Average fill-in (lower is better) on $\textsc{99.graph}$, comparing masking vs.\ no masking during training.}
    \label{99Graphablation}
\end{figure}

\begin{lstlisting}[language=bash]
  $ python main.py datasets/100.graph --output_file non_masking_results/100.graph  
  --policy_sizes 64 32 --total_timesteps 1_000_000 --learning_rate 0.0001   
  --parallel_envs 10 --node_dim 16 --ent_coef 0.002 --action_masking 0
\end{lstlisting}

\begin{figure}[H]
    \centering
    \includegraphics[width=0.7\linewidth]{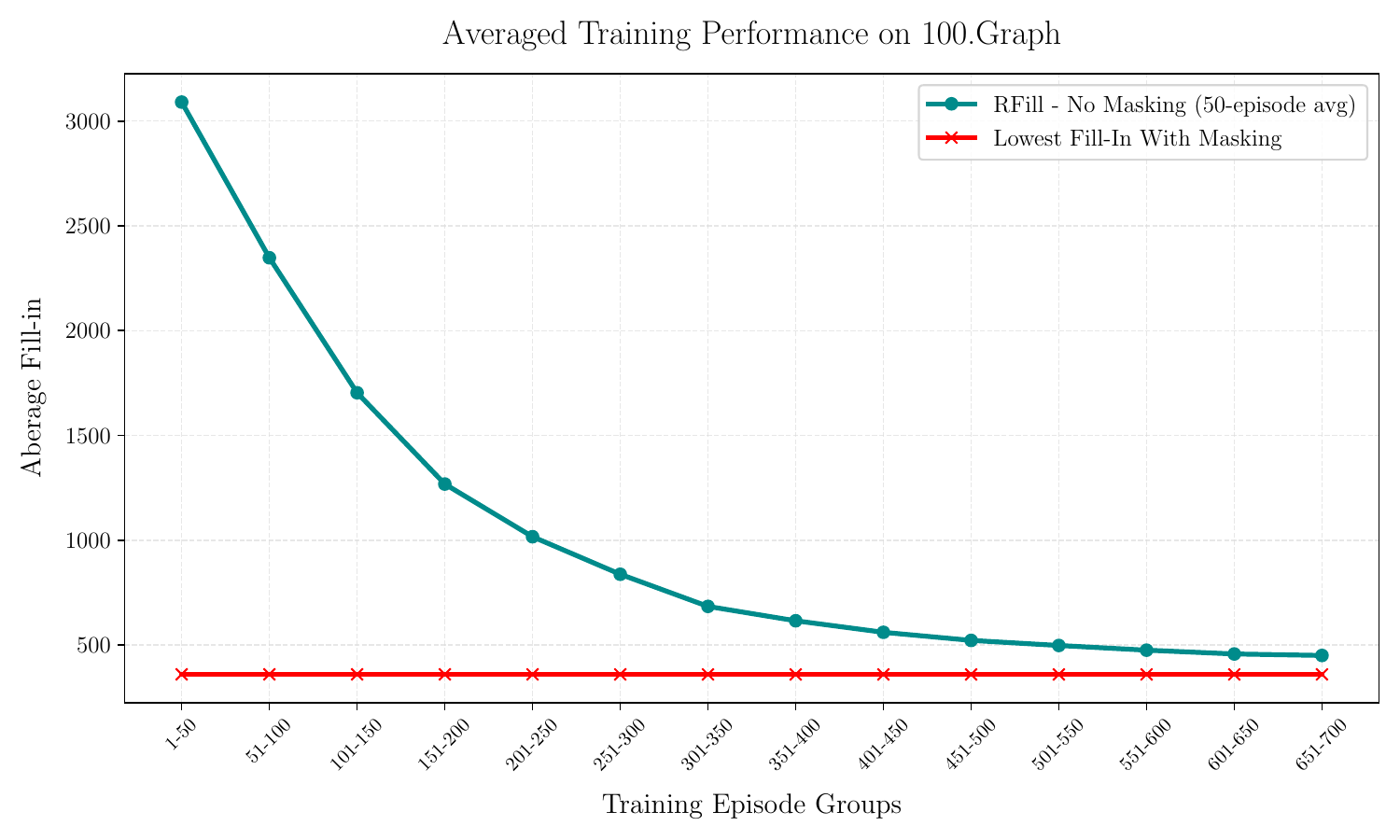}
    \caption{Average fill-in (lower is better) on $\textsc{100.graph}$, comparing masking vs.\ no masking during training.}
    \label{100Graphablation}
\end{figure}

\clearpage
\section{Generalization Experiment}
\label{genexp}

For this experiment, we sample $35$ graphs from $G(n, p), n=50, p=0.2$ using the following hyperparameters. 

\begin{lstlisting}[language=bash]
  $ python gnp.py  --output_file non_masking_results/gnp.graph   
  --policy_sizes  --total_timesteps 500_000 --learning_rate 0.0001 
  --parallel_envs 35  --node_dim 16  --action_masking 1 --ent_coef 0.01
\end{lstlisting}

The model is trained on all $35$ environments in parallel using action masking for $500,000$ timesteps. Next, we sample $200$ new evaluation graph instances from $G(n, p), n=50, p=0.2$ and evaluate the fill-in from following the trained policy for each instance. To do this, we sample $25$ elimination orders for each graph using the trained model, and report the minimum fill-in order found for each graph. 

\begin{figure}[H]
    \centering
    \includegraphics[width=0.6\linewidth]{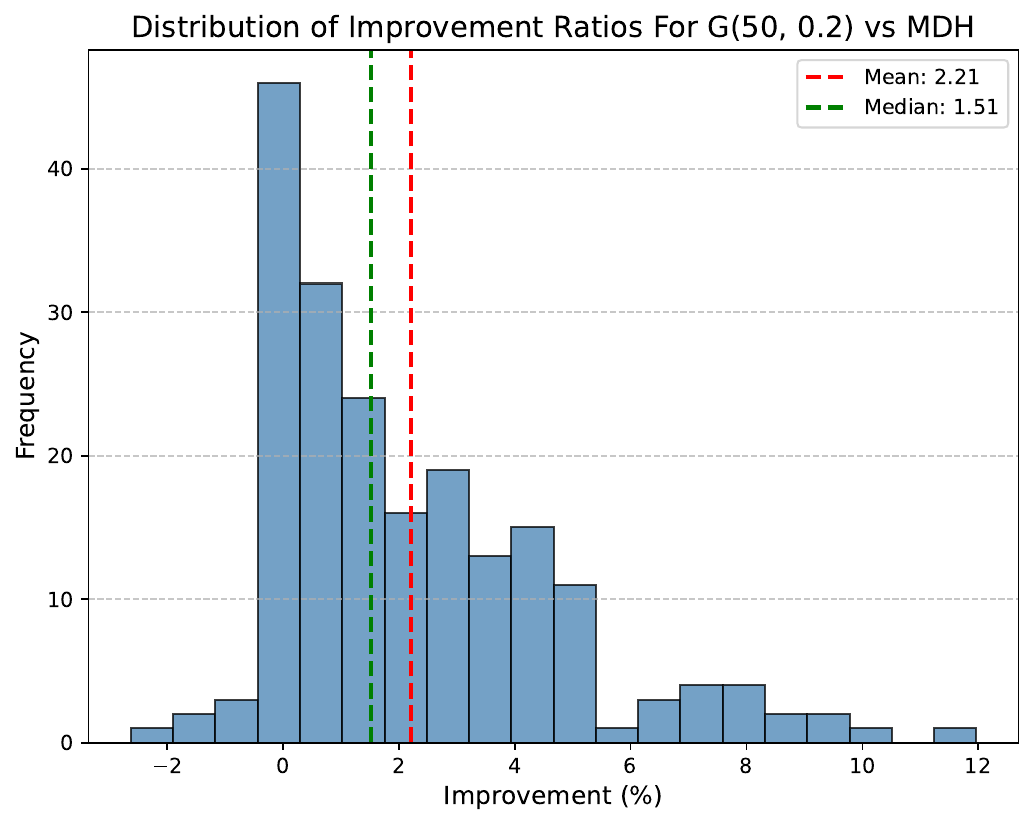}
    \caption{Improvement percentages over the $200$ instances of $G(50, 0.2)$ evaluation graphs compared to $\textsc{MDH}$}
    \label{improvementsmdh}
\end{figure}

\begin{figure}[H]
    \centering
    \includegraphics[width=0.6\linewidth]{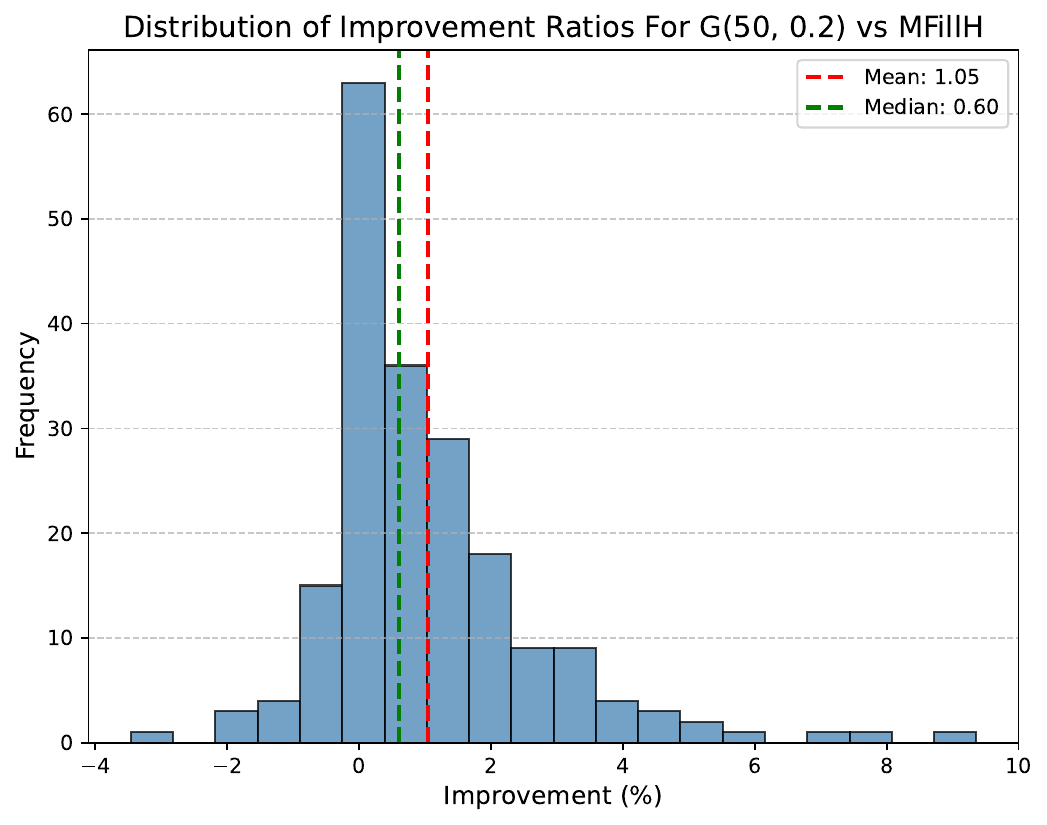}
    \caption{Improvement percentages over the $200$ instances of $G(50, 0.2)$ evaluation graphs compared to $\textsc{MFillH}$}
    \label{improvementsmfillh}
\end{figure}

\begin{figure}[H]
    \centering
    \includegraphics[width=0.6\linewidth]{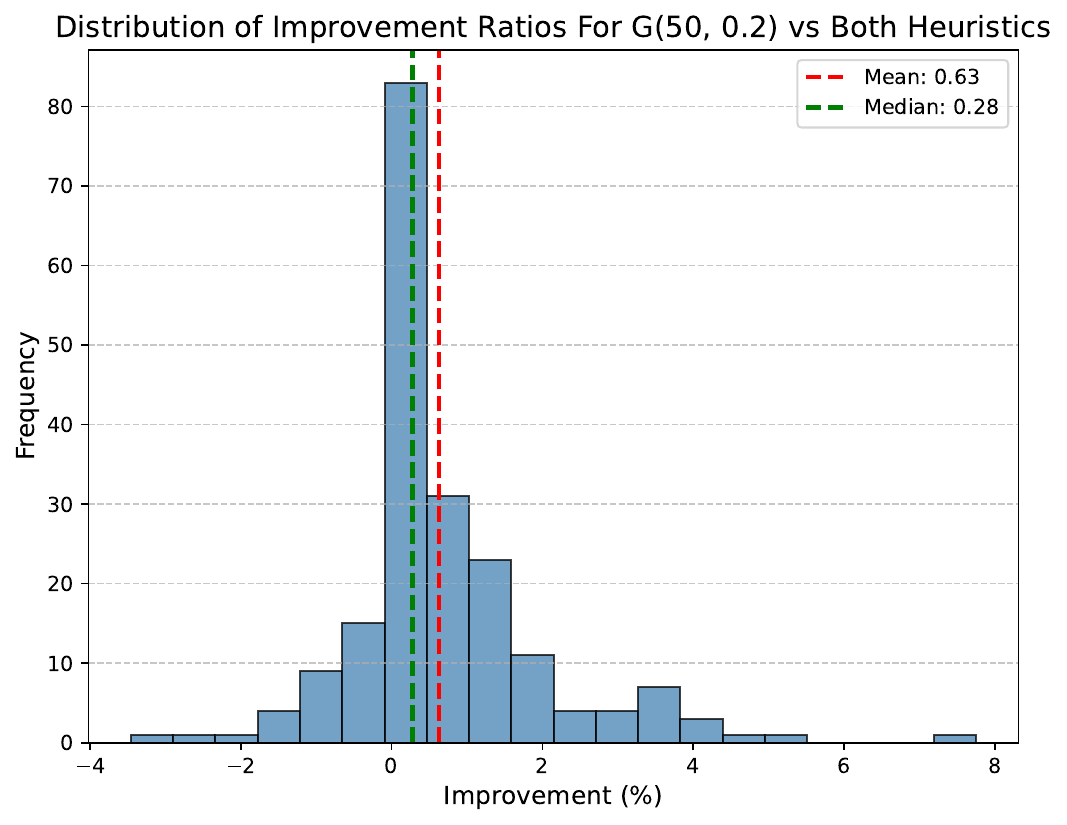}
    \caption{Improvement percentages over the $200$ instances of $G(50, 0.2)$ evaluation graphs compared to both $\textsc{MDH}$ and $\textsc{MFillH}$}
    \label{improvementsboth}
\end{figure}

\cref{improvementsmdh} shows the improvements of the trained policy vs the \textsc{MDH} heuristic. \cref{improvementsmfillh} shows the improvements of the trained policy vs the \textsc{MFillH} heuristic. Finally, \cref{improvementsboth} shows the improvements of the trained policy vs both heuristics (i.e. compared to the minimum of both heuristics). 

On average, the learned heuristic generalizes and gives an elimination order than is on average a 2.21\% improvement over $\textsc{MDH}$, a 1.05\% improvement over \textsc{MFillH}, and 0.63\% over both heuristics. However, it does not always lead to a better fill-in for all graphs. Hence, the learned heuristic can be bootstrapped with the other heuristics to generate strictly better heuristics. 

\end{document}